\documentclass{article}

\usepackage{microtype}
\usepackage{graphicx}
\usepackage{subcaption}
\usepackage{booktabs} % for professional tables

\usepackage{hyperref}

\usepackage[preprint]{icml2026}

\usepackage{amsmath}
\usepackage{amssymb}
\usepackage{mathtools}
\usepackage{amsthm}

\usepackage[capitalize,noabbrev]{cleveref}

\usepackage{siunitx}
\usepackage{pifont}
\usepackage{amssymb}
\usepackage{booktabs}
\usepackage{makecell}
\usepackage{multirow}
\usepackage{enumitem}
\usepackage{tabularx}
\usepackage{booktabs}
\usepackage{amsfonts}

\theoremstyle{plain}

\theoremstyle{definition}

\theoremstyle{remark}

\usepackage[textsize=tiny]{todonotes}

\AtBeginDocument{
  \setlength{\abovedisplayskip}{7pt}
  \setlength{\belowdisplayskip}{7pt}
  \setlength{\abovedisplayshortskip}{3pt}
  \setlength{\belowdisplayshortskip}{3pt}
}
\icmltitlerunning{Rethinking EEG Foundation Model Pretraining}

\begin{document}

\twocolumn[
\icmltitle{Standing on the Shoulders of Giants: \\ Rethinking EEG Foundation Model Pretraining via Multi-Teacher Distillation}
  % \icmltitle{Rethinking EEG Foundation Model Pretraining}



  \icmlsetsymbol{equal}{*}

  \begin{icmlauthorlist}
    \icmlauthor{Chenqi Li}{yyy}
    \icmlauthor{Yu Liu}{yyy}
    \icmlauthor{Shuo Zhang}{yyy}
    \icmlauthor{Timothy Denison}{yyy}
    \icmlauthor{Tingting Zhu}{yyy}
  \end{icmlauthorlist}

  \icmlaffiliation{yyy}{Department of Engineering Science, University of Oxford}

  \icmlcorrespondingauthor{Chenqi Li}{chenqi.li@eng.ox.ac.uk}

  \icmlkeywords{Machine Learning, ICML}

  \vskip 0.3in
]



\printAffiliationsAndNotice{}  % no special notice (required even if empty)

\begin{abstract}
Pretraining for electroencephalogram (EEG) foundation models has predominantly relied on self-supervised masked reconstruction, a paradigm largely adapted from and inspired by the success of vision and language foundation models. However, unlike images and text, EEG datasets are notoriously expensive to collect and characterized by low signal-to-noise ratio. These challenges introduce difficulties in scaling the EEG foundation models and capturing the underlying neural semantics through reconstruction. In this work, we ask the question: can we stand on the shoulders of well-established foundation models from well-represented modalities to bootstrap the pretraining of EEG foundation models? We first demonstrate that mainstream foundation models, such as those from vision and time series, transfer surprisingly well to EEG domain. To this end, we propose the Multi-Teacher Distillation Pretraining (MTDP) framework for pretraining EEG foundation models via a two-stage multi-teacher distillation. In the first stage, we introduce a learnable gating network to fuse representations from diverse teachers (e.g., DINOv3 and Chronos) via a masked latent denoising objective. In the second stage, we distill the fused representation into an EEG foundation model. Extensive evaluations across 9 downstream tasks and 12 datasets demonstrate that our MTDP-based EEG foundation model outperforms its self-supervised counterparts while requiring only 25\% of the pretraining data.
\end{abstract}

\section{Introduction}
Electroencephalogram (EEG) signals record the spontaneous electrical activity of the brain through electrodes placed on the scalp  \cite{siuly2017electroencephalogram}. It has become a fundamental tool for assisting clinical diagnosis, advancing neuroscience and cognitive science research, developing brain-computer interface (BCI), and is considered the gold standard for sleep staging \cite{berry2017aasm} and epilepsy diagnosis \cite{fisher2017operational}. Although EEG recordings have traditionally been analyzed by highly trained specialists, deep learning has not only succeeded in automating such analysis, but also in unlocking new applications for EEG where data were not traditionally read by humans. Deep learning models, such as convolutional neural networks and recurrent neural networks, have demonstrated exceptional performance across a variety of tasks: seizure detection \cite{alotaiby2014eeg}, sleep staging \cite{aboalayon2016sleep}, emotion recognition \cite{li2022eeg}, and motor imagery classification \cite{altaheri2023deep}. However, early EEG-based deep learning models are often tailored to specific tasks and datasets, with limited ability to generalize beyond strictly defined experimental setups.

% Compared to other methods for studying brain functions (such as functional magnetic resonance imaging (fMRI), positron emission tomography (PET), and electrocorticography (ECoG)), the main advantages of EEG include low hardware cost, portability of the device,  high temporal resolution, and a high tolerance of subject movement. Despite its limited spatial resolution,

Following the remarkable success of foundation models (FMs) in computer vision and natural language processing, there is growing interest in developing similar FMs for EEG. This presents a paradigm shift away from training task and dataset specific models, to learning universal neural representations for models that can be adapted to a wide range of downstream tasks \cite{yang2023biot, zhang2023brant}. In particular, the pretraining of existing EEG FMs predominantly relies on masked reconstruction, where the model learns to predict missing patches of the input signal. Subsequently, these pretrained models are fine-tuned with a classification head, demonstrating state-of-the-art performance across diverse benchmarks \cite{jiang2024large, wang2025cbramod, ouahidi2025reve}.

Unlike natural language or images, EEG data are significantly more challenging to collect and share due to strict privacy constraints. Consequently, existing EEG datasets remain orders of magnitude smaller than the internet-scale corpora used to train vision \cite{simeoni2025dinov3} and language \cite{grattafiori2024llama} FMs, thereby challenging the scaling of EEG FMs. Furthermore, EEG signals are characterized by an exceptionally low signal-to-noise ratio, by which reconstruction-based objectives inherently prioritize the modelling of artifacts and noise, effectively directing the model away from meaningful neural dynamics. The scarcity of data and the sensitivity to noise call for more data-efficient representation learning strategies that leverage external priors, rather than rely solely on masked reconstruction. This naturally raises the question: can we stand on the shoulders of well-established FMs from well-represented modalities to bootstrap the pretraining of an EEG FM?

% Although EEG FMs present a significant leap forward in EEG analysis, they possess weaker universal capabilities compared to established counterparts in computer vision and natural language processing. Vision and language FMs are capable of extracting generalizable representations that can be utilized via linear probing to achieve high performance on downstream tasks \cite{simeoni2025dinov3}. In contrast, most existing EEG FMs rely on fine-tuning for every downstream application \cite{wang2025cbramod}. 

\begin{figure}
    \centering
    \includegraphics[width=\linewidth]{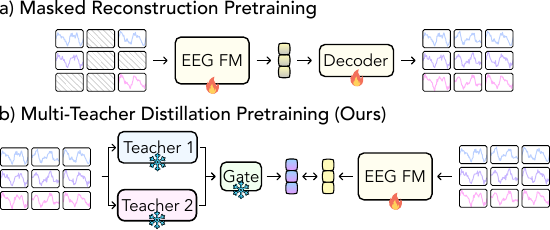}
    \caption{Comparison of EEG Foundation Model Pretraining. a) The conventional self-supervised pretraining where EEG foundation model reconstructs missing patches in the temporal, frequency or latent-domain. b) The proposed framework to bootstrap EEG foundation model pretraining by standing on the shoulders of well-established foundation models from well-represented modalities.}
    \label{fig:methodscompare}
\end{figure}

In this work, we start by showing that mainstream FMs are surprisingly good at extracting EEG representations through a series of linear probing experiments. We then propose the Multi-Teacher Distillation Pretraining (MTDP) framework to pretrain EEG FM via two-stage multi-teacher distillation. In the first stage, we introduce a learnable gating network to weigh the importance of each teacher model and fuse teacher representations in an unsupervised manner. The second stage pretrains EEG FM by learning from the fused teacher representations through distillation. Figure \ref{fig:methodscompare} provides a comparison of masked reconstruction pretraining used in existing EEG FMs and the proposed MTDP framework. Our contributions can be summarized as follows:
\begin{itemize}[leftmargin=10px, itemsep=5pt, topsep=0pt, parsep=0pt]
    \item We show that vision FM, DINOv3, despite being trained on a different modality, transfer surprisingly well to EEG tasks, outperforming specialized EEG FMs under the linear probing setting.
    \item We explore multi-teacher knowledge distillation from vision and time series FMs as a data-efficient alternative to large-scale self-supervised pretraining for EEG.
    \item To leverage the complementary representations from teacher FMs of different modalities, we present a learnable gating network to weigh each teacher via a masked latent denoising objective and fuse teacher representations in an unsupervised manner.
    \item We demonstrate that distilled EEG FM consistently surpasses state-of-the-art self-supervised counterparts, in both linear probing and full fine-tuning settings, while only requiring ~25\%\ of pretraining data.
\end{itemize}

\section{Related Works}
% \subsection{Related Works}
\subsection{EEG Foundation Model}
Deep learning models for automated EEG analysis have undergone a paradigm shift from task-specific supervised learning to FMs that are capable of achieving state-of-the-art results on a variety of downstream tasks. Early efforts demonstrated the viability of learning universal biosignal representations by aggregating diverse datasets for pretraining \cite{yang2023biot}, followed by subsequent work that scaled the pretraining to larger datasets \cite{zhang2023brant, yuan2024brant, jiang2024large}. The pretraining strategy for EEG FMs focuses on self-supervised learning, specifically masked reconstruction, aiming to reconstruct masked segments in the temporal domain \cite{yang2023biot, chen2024eegformer, wang2025cbramod}, frequency domain \cite{jiang2024large}, or latent representation spaces \cite{cui2024neuro, jiang2024large}. Concurrently, research has focused on architectural designs to handle the heterogeneous nature of EEG electrode configurations by introducing topology-agnostic mechanisms to generalize across arbitrary electrode montages \cite{ouahidi2025reve, fang2025neuript}, to improve scaling by exploring efficient state-models \cite{doner2025luna, tegon2025femba}, or to incorporate joint learning with additional modalities such as iEEG \cite{yuan2025brainwave}, MEG \cite{xiao2025brainomni}, and other biosignals \cite{luo2025foundation}. However, despite these advances in scaling and architecture, alternative pretraining objectives to masked reconstruction remain underexplored \cite{liu2026eegfoundationmodelsprogresses}. 

\subsection{Repurposing Models for Time Series}
The success of large-scale pretraining in natural language processing and computer vision has inspired a growing interest in repurposing or reprogramming existing FMs for time series \cite{li2023time, jin2024timea, zhong2025time}, rather than training models from scratch. Early efforts investigated the feasibility of whether the structural understanding of vision transformers and the reasoning and sequential capabilities of LLMs could be transferred to time series applications. It was demonstrated that plotting irregularly sampled time series into line graph images allows pretrained vision transformers to outperform specialized baselines, offering a robust and simplified framework for extreme data scarcity \cite{li2023time}. Time-LLM \cite{jin2024timea} demonstrated that a reprogramming layer that aligns time series patches with text prototype embeddings enables LLMs to perform forecasting without changing the backbone parameters. Time-VLM \cite{zhong2025time} further extended the repurposing paradigm to vision-language models and established a new frontier for multimodal time series forecasting. Similar cross-modal repurposing has also been adopted for domain-specific applications, such as medical time series classification \cite{huang2024repurposing}, zero-shot ECG learning \cite{li2024frozen}, and prediction of healthcare events \cite{yu2025healthcare}. These studies highlight the untapped potential of cross-modal priors for EEG representation learning, offering a promising yet unexplored path for their integration into EEG FM pretraining.

\subsection{Multi-Teacher Distillation}
The concept of transferring knowledge from one model to another was formalized by \cite{hinton2015distilling}. Although single-teacher distillation is effective, multi-teacher distillation seeks to synthesize diverse perspectives from different teacher models, through collaborative frameworks \cite{pham2022collaborative} or dynamic weighing of teacher contributions \cite{liu2020adaptive}. Recent literature has increasingly explored distillation from Mixture-of-Experts (MoE) architectures \cite{xie2024mode, al2025mixture, kim2025every, li2024multi}. These systems can be viewed as an extension of multi-teacher frameworks, wherein a gating mechanism dynamically selects the most relevant teachers. A notable application is AM-RADIO \cite{ranzinger2024radio}, which employs multi-teacher distillation to amalgamate the unique strengths of each teacher into a unified model. We adopt a similar philosophy by utilizing a learnable gate to fuse cross-modal priors, effectively treating frozen mainstream FMs as domain-specific teachers.

\section{Preliminary Experiment} \label{sec:preliminaryexperiments}
% \subsubsection{Comparing the Representation of EEG Foundation Model and Vision Foundation Model}
To evaluate the quality of representation of pretrained EEG FMs versus mainstream FMs, we perform a linear probing analysis comparing the state-of-the-art EEG FM CBraMod \cite{wang2025cbramod} against the vision FM DINOv3 \cite{simeoni2025dinov3}. We extract representations from EEG signals using both models as frozen feature extractors. The cross-modal adaptation protocol used to project EEG signals into a format compatible with DINOv3 is detailed in Section \ref{sec:eegpreprocessing}. Finally, we evaluate the linear separability of these representations by training a logistic regression classifier that serves as a proxy for the quality of the captured neural dynamics. By restricting the classifier to a linear architecture, we ensure that the performance reflects the intrinsic quality of the representations rather than the capacity of the downstream head to compensate for poor representations.

\begin{figure}[!ht]
    \centering
    \includegraphics[width=\linewidth]{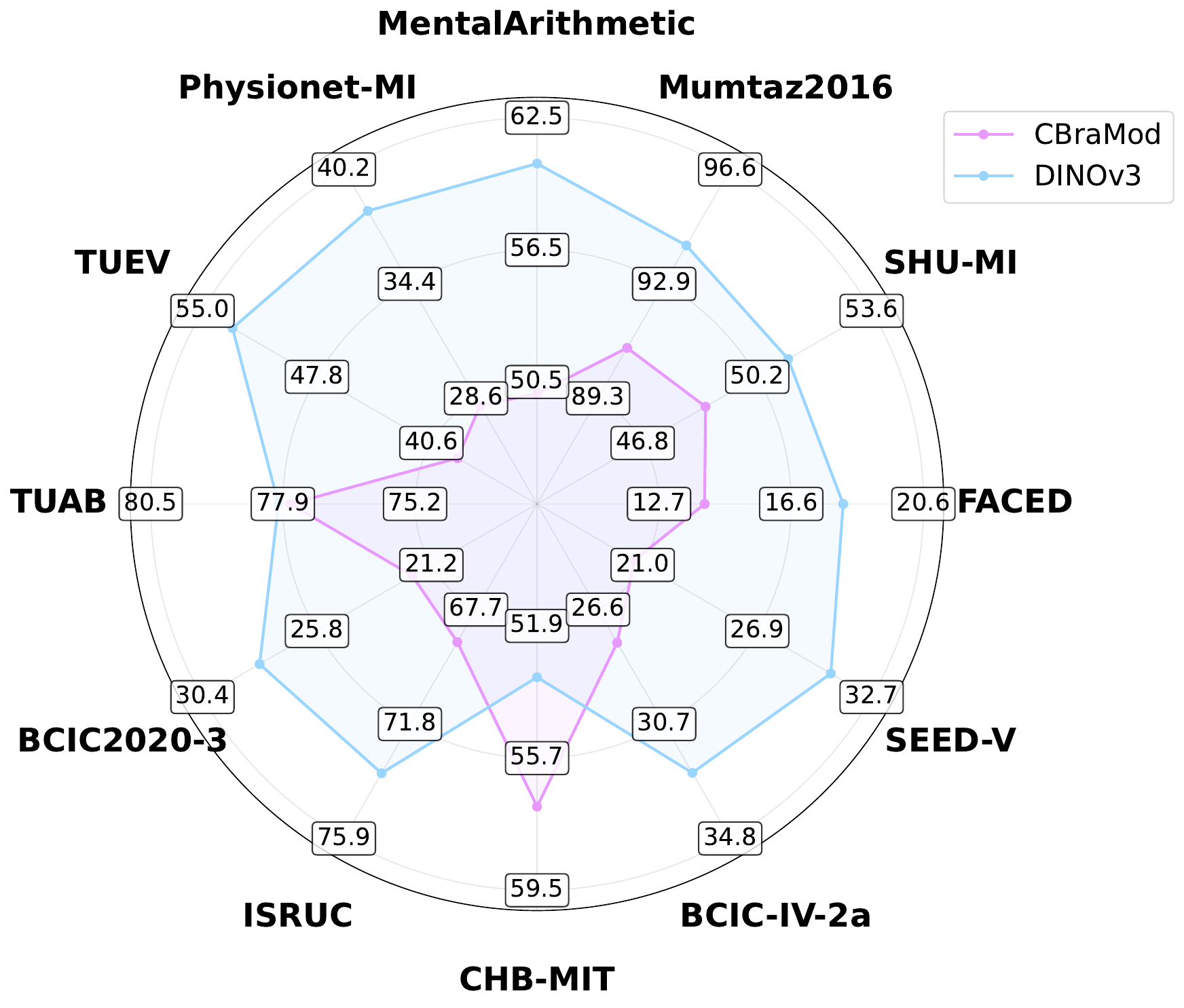}
    \caption{Linear probing performance of CBraMod and DINOv3 on EEG downstream tasks. Balanced Accuracy (\%).}
    \label{fig:linearprobingdinocbramod}
\end{figure}

Figure \ref{fig:linearprobingdinocbramod} compares the performance of the linear probing in a variety of downstream EEG tasks. DINOv3 consistently outperforms CBraMod across all datasets except CHB-MIT, suggesting that, compared to EEG FMs, large-scale vision FMs generalize well to EEG tasks. These findings provide strong motivation to reconsider the traditional masked reconstruction pretraining for EEG FMs, and instead explore leveraging the knowledge from mainstream FMs.

\section{Methodology}

\subsection{Problem Definition}
Let $\mathcal{D} = \{ \mathbf{x}_i \}_{i=1}^N$ denote an unlabeled EEG dataset containing $N$ samples. Each sample $\mathbf{x} \in \mathbb{R}^{C \times T}$ represents a multi-channel recording with $C$ channels across $T$ time steps. We define an EEG FM $f_\theta: \mathbb{R}^{C \times T} \rightarrow \mathbb{R}^{d}$, parameterized by $\theta$, which serves as an encoder that maps the EEG sample $\mathbf{x}$ into a $d$-dimensional latent representation $\mathbf{h} \in \mathbb{R}^{d}$. 

The objective of EEG FM pretraining is to optimize the parameters $\theta$ to maximize generalization performance across a set of $N_{\text{task}}$ downstream tasks $\mathcal{T} = \{\mathcal{T}_t\}_{t=1}^{N_{\text{task}}}$. Each task $\mathcal{T}_t$ is associated with a dataset $\mathcal{D}_t$ partitioned into training $\mathcal{D}_{t}^{\text{train}}$, validation $\mathcal{D}_{t}^{\text{val}}$, and test $\mathcal{D}_{t}^{\text{test}}$ subsets. The training set consists of $N_\text{train}$ EEG-label pairs, $\mathcal{D}_{t}^{\text{train}} = \{(\mathbf{x}_i, y_i)\}_{i=1}^{N_\text{train}}$, with the validation and test sets defined analogously. 
The pretraining optimization objective is formulated as follows:
\begin{equation}
    \theta^* = \arg \max_{\theta} \mathbb{E}_{\mathcal{T}_t \in \mathcal{T}} \left[ \mathcal{P}_{t} (h_{\eta_t} \circ f_{\theta_t}, \mathcal{D}_{t}^{\text{test}}) \right]
\end{equation}
where $\mathcal{P}_{t}$ denotes the performance metric (e.g., accuracy or F1-score) for the task $\mathcal{T}_t$, and $h_{\eta_t}$ is a task-specific head parameterized by $\eta_t$. The operator $\circ$ denotes function composition. The adapted parameters $\theta_t$ and $\eta_t$ are derived from the pretrained initialization $\theta$ and random initialization $\eta$ via an adaptation algorithm $\mathbb{A}$:
\begin{equation}
(\theta_t, \eta_t) = \mathbb{A}(\theta, \eta, \mathcal{D}_{t}^{\text{train}}, \mathcal{D}_{t}^{\text{val}})
\end{equation}
The algorithm $\mathbb{A}$ typically corresponds to linear probing, where $\theta_t = \theta$, or fine-tuning. Crucially, the pretraining phase lacks prior knowledge of $\mathcal{T}$, requiring the encoder $f_\theta$ to learn universal representations of neural dynamics agnostic to tasks.

\subsection{Two-Stage Multi-Teacher Distillation Pretraining} \label{sec:twostagedistillation}
The core idea of the proposed MTDP framework is to pretrain EEG FMs by leveraging the knowledge from mainstream FMs. We achieve this by injecting the knowledge of well-established FMs from well-represented modalities (teachers) into an EEG FM (student) via a specialized distillation process. The choice of EEG FM architecture is flexible, which can be an existing EEG FM like CBraMod.

\begin{figure*}[!ht]
    \centering
    \includegraphics[width=\linewidth]{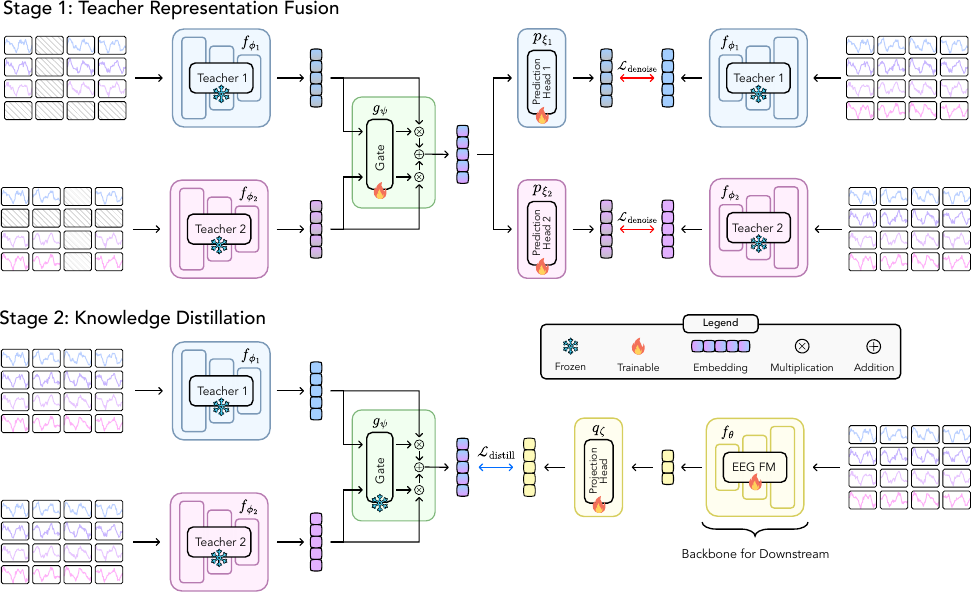}
    \caption{Overview of Two-Stage Multi-Teacher Distillation Pretraining (MTDP). Stage 1: Teacher Representation Fusion. A learnable gating network is introduced to weigh and fuse representations from frozen teacher models. The gate is trained via a masked latent denoising objective. Stage 2: Knowledge Distillation. The fused teacher representation acts as the target to pretrain the student EEG foundation model. The distillation loss is minimized to align the student representations with the fused representations.}
    \label{fig:method}
\end{figure*}

\begin{algorithm}[tb]
\caption{MTDP Framework: Two-Stage Multi-Teacher Distillation Pretraining}
\label{alg:distillation}
\begin{algorithmic}
   \STATE {\bfseries Input:} \hspace{0.15cm} Pretraining dataset $\mathcal{D}$ \\
                             \hspace{1.25cm} Pretrained teachers $\{f_{\phi_k}\}_{k=1}^K$
   \STATE {\bfseries Output:} EEG foundation model $f_\theta$ \\
   $\triangleright$ \textbf{Stage 1: Teacher Representation Fusion}
   \STATE Initialize gating network $g_\psi$
   \STATE Initialize prediction heads $\{p_{\xi_k}\}_{k=1}^K$
   \FOR{iteration $\leftarrow$ 1 to $N_{\text{iter}}$}
      \STATE Sample EEG $\mathbf{x} \sim \mathcal{D}$
      \STATE Apply mask $\tilde{\mathbf{x}} \leftarrow \mathbf{x} \odot \mathbf{m}$
      \FOR{$k$ $\leftarrow$ 1 to $K$}
         \STATE Compute $\tilde{\mathbf{h}}_k = f_{\phi_k}(\tilde{\mathbf{x}})$ and $\mathbf{h}_k = f_{\phi_k}(\mathbf{x})$
      \ENDFOR
      \STATE Compute $\mathbf{w} = [w_1, ..., w_K] = g_\psi([\tilde{\mathbf{h}}_1; \dots; \tilde{\mathbf{h}}_K])$
      \STATE Compute $\tilde{\mathbf{h}}_{\text{fused}} = \sum_{k=1}^K w_k \cdot \tilde{\mathbf{h}}_k$
      \STATE Compute $\mathcal{L}_{\text{denoise}} = \sum_{k=1}^{K} \|p_{\xi_k}(\tilde{\mathbf{h}}_{\text{fused}}) - \mathbf{h}_{k}\|^2_2$
      \STATE Update $\psi, \xi_k$ using $\nabla_\psi\mathcal{L}_{\text{denoise}}$, $\nabla_{\xi_k}\mathcal{L}_{\text{denoise}}$
   \ENDFOR
   
   $\triangleright$ \textbf{Stage 2: Knowledge Distillation}
   \STATE Initialize EEG foundation model $f_\theta$ 
   \STATE Initialize projection heads $\{q_{\zeta}\}_{k=1}^K$
   \FOR{iteration $\leftarrow$ 1 to $N_{\text{iter}}$}
      \STATE Sample EEG $\mathbf{x} \sim \mathcal{D}$
      \STATE Compute $\mathbf{h}_{\text{fused}} = \sum_{k=1}^K w_k \cdot f_{\phi_k}(\mathbf{x})$
      \STATE Compute $\mathcal{L}_{\text{distill}} = 1 - \frac{\langle q_{\zeta} \circ f_{\theta}(x), \mathbf{h}_{\text{fused}} \rangle}{||q_{\zeta} \circ f_{\theta}(x)||_2 \cdot ||\mathbf{h}_{\text{fused}}||_2}$
      \STATE Update $\theta, \zeta$ using $\nabla_\theta\mathcal{L}_{\text{distill}}, \nabla_\zeta\mathcal{L}_{\text{distill}}$
   \ENDFOR
\end{algorithmic}
\end{algorithm}

EEG signals are high-dimensional and non-stationary, characterized by a complex interplay of spatial, temporal, and spectral components that reflect the underlying neurophysiological activities. Although Section \ref{sec:preliminaryexperiments} shows that vision FMs have demonstrated promising potential to extract useful EEG representations, a single teacher often lacks the breadth to capture all of the aforementioned dimensions simultaneously. To address this, our MTDP framework seeks to synthesize complementary insights from FMs of different modalities that serve to guide the EEG model toward a holistic understanding of the multifaceted nature of EEG more effectively than any single teacher could alone.

To leverage the complementary representations from different modalities, our MTDP framework has two stages: 
\begin{itemize}[leftmargin=10px, itemsep=5pt, topsep=0pt, parsep=0pt]
    \item \textbf{Stage 1: Teacher Representation Fusion.} We introduce a learnable gating network to weigh the importance of each teacher model using a masked latent denoising objective. The goal is to synthesize a unified teacher representation in an unsupervised manner.
    \item \textbf{Stage 2: Knowledge Distillation.} The unified teacher representation serves as the distillation target for a randomly initialized EEG FM student. The goal is to transfer the synthesized knowledge from different teachers into a compact EEG FM.
\end{itemize}
An overview of the method is illustrated in Figure \ref{fig:method}.

\subsubsection{Stage 1: Teacher Representation Fusion} \label{sec:stage1pretrain}
The main challenge in the fusion of teacher representation is to identify the relative importance of each teacher in an unsupervised manner. Although teachers can be weighted using task-specific objective functions, such functions are unavailable during unsupervised pretraining when downstream tasks are unknown. To address this issue, we design a masked latent denoising objective to automatically weigh each teacher. Intuitively, the objective encourages the fusion of masked teacher representations such that it can best predict each of the unmasked teacher representations.

Given an EEG sample $\mathbf{x}$ from the pretraining dataset $\mathcal{D}$, we construct a masked version $\tilde{\mathbf{x}}$ by applying a binary mask $\mathbf{m} \in \{0, 1\}^{C \times T}$:
\begin{equation}
\tilde{\mathbf{x}} = \mathbf{x} \odot \mathbf{m},
\end{equation}
where $\odot$ denotes element-wise product. We apply two types of masking strategies:
\begin{itemize}[leftmargin=10px, itemsep=5pt, topsep=0pt, parsep=0pt]
    \item \textbf{Segment Masking}: A contiguous window between one to two seconds is masked across all channels to encourage the model to learn temporal dynamics.
    \item \textbf{Channel Dropout}: A random channel is dropped for all timesteps to capture spatial correlation and redundancy.
\end{itemize} 

Let $\{f_{\phi_k}\}_{k=1}^K$ be a set of $K$ pretrained teacher models, where each model is parametrized by $\phi_k$. For each teacher $k$, we extract the representation from the unmasked and masked EEG samples:
\begin{equation} \label{eq:eegtorepresentation}
    \mathbf{h}_{k} = f_{\phi_k}(\mathbf{x}) \quad \text{and} \quad \tilde{\mathbf{h}}_{k} = f_{\phi_k}(\tilde{\mathbf{x}})
\end{equation}
Let $g_\psi$ be the gating network parametrized by $\psi$. The gate takes as input the concatenated masked teacher representations and outputs a scalar weight for each teacher:
\begin{equation} \label{eq:gatingnetwork}
\mathbf{w} = [w_{1}, w_{2}, \dots, w_{K}] = g_\psi([\tilde{\mathbf{h}}_{1}; \tilde{\mathbf{h}}_{2}; \dots; \tilde{\mathbf{h}}_{K}])
\end{equation}
The fused teacher representation is computed as the weighted sum of the teacher representations:
\begin{equation} \label{eq:fuserepresentation}
    \tilde{\mathbf{h}}_{\text{fused}} = \sum_{k=1}^{K} w_{k} \cdot \tilde{\mathbf{h}}_{k}
\end{equation}

The fused representation $\tilde{\mathbf{h}}_{\text{fused}}$ amalgamates the representations from different teachers and will serve as the distillation target for stage 2, facilitating the transfer of knowledge from multiple teachers into an EEG FM student. The gating network is trained using the masked latent denoising objective:
\begin{equation}
\mathcal{L}_{\text{denoise}} = \sum_{k=1}^{K} \|p_{\xi_k}(\tilde{\mathbf{h}}_{\text{fused}}) - \mathbf{h}_{k}\|^2_2,
\end{equation}
where $\{p_{\xi_k}\}_{k=1}^K$ is a set of prediction heads parametrized by $\xi_k$ for each teacher.

\subsubsection{Stage 2 Knowledge Distillation} \label{sec:stage2pretrain}
With the pretrained gating network, the objective of stage 2 is to distill the fused teacher representation into the student model. For an EEG sample $\mathbf{x}$ from the pretraining dataset, we compute the representation from each teacher $\mathbf{h}_{k} = f_{\phi_k}(\mathbf{x})$ and the weight for each teacher $\mathbf{w} = [w_{1}, \dots, w_{K}] = g_\psi([\mathbf{h}_{1}; \mathbf{h}_{2}; \dots; \mathbf{h}_{K}])$. The fused representation is computed as:
\begin{equation}
    \mathbf{h}_{\text{fused}} = \sum_{k=1}^{K} w_{k} \cdot \mathbf{h}_{k}
\end{equation}
To train the student model, we minimize the distillation loss as the cosine similarity between the student representation and the target fused teacher representation:
\begin{equation}
    \mathcal{L}_{\text{distill}} = 1 - \frac{\langle q_{\zeta} \circ f_{\theta}(x), \mathbf{h}_{\text{fused}} \rangle}{||q_{\zeta} \circ f_{\theta}(x)||_2 \cdot ||\mathbf{h}_{\text{fused}}||_2},
\end{equation}
where $q_{\zeta}$ is a learnable projection head parameterized by ${\zeta}$ to match the dimensionality of the student representation with that of the fused teacher representation. $\langle \cdot, \cdot \rangle$ denotes the inner product operator. For downstream fine-tuning, teacher models $\{f_{\phi_k}\}_{k=1}^K$, gating network $g_\psi$, prediction heads $\{p_{\xi_k}\}_{k=1}^K$, and projection head $q_{\zeta}$ are discarded.

A detailed description of the MTDP Framework is provided in Algorithm \ref{alg:distillation}.

\section{Experiments}
\begin{table*}[!h]
\centering
\caption{Overview of Downstream Datasets}
\label{tab:downstreamdatasets}
\resizebox{0.95\textwidth}{!}{
    \begin{tabular}{lcccccc}
    \toprule
    \textbf{Task} & \textbf{Dataset} & \textbf{Rate} & \textbf{\# Channels} & \textbf{Duration} & \textbf{\# Samples} & \textbf{Label} \\ \midrule
    Motor Imagery Classification & PhysioNet-MI & 160 \unit{\hertz} & 64 & 4s & 9,837 & 4-class \\
     & SHU-MI & 250 \unit{\hertz} & 32 & 4s & 11,988 & 2-class \\
     & BCIC-IV-2a & 250 \unit{\hertz} & 22 & 4s & 5088 & 4-class \\
    Emotion Recognition & FACED & 250 \unit{\hertz} & 32 & 10s & 10,332 & 9-class \\
     & SEED-V & 1000 \unit{\hertz} & 62 & 1s & 115,001 & 5-class \\
    Sleep Staging & ISRUC & 200 \unit{\hertz} & 6 & 30s & 90,187 & 5-class \\
    Seizure Detection & CHB-MIT & 256 \unit{\hertz} & 16 & 10s & 326,993 & 2-class \\
    Imagined Speech Classification & BCIC2020-3 & 256 \unit{\hertz} & 64 & 3s & 6,000 & 5-class \\
    Mental Disorder Diagnosis & Mumtaz2016 & 256 \unit{\hertz} & 19 & 5s & 7,143 & 2-class \\
    Mental Stress Detection & MentalArithmetic & 500 \unit{\hertz} & 20 & 5s & 1,707 & 2-class \\
    Event Type Classification & TUEV & 250 \unit{\hertz} & 16 & 5s & 113,353 & 6-class \\
    Abnormal Detection & TUAB & 250 \unit{\hertz} & 16 & 10s & 409,455 & 2-class \\ 
    \bottomrule
    \end{tabular}
}
\end{table*}

\begin{table*}[!ht]
\centering
\caption{Comparison of EEG foundation model performance on 12 downstream tasks. ``25\%'' and ``100\%'' indicate pretraining dataset size. Metrics acronyms:  Balanced Accuracy (B. Acc.), Weighted F1 (F1-W), Cohen's Kappa (Kappa).}
\label{tab:eegfmresults}
\small
\setlength{\tabcolsep}{3.5pt}
\renewcommand{\arraystretch}{1.15}
\begin{tabular}{l | c | ccc | ccc | ccc}
\toprule[1.5pt]
\multirow{2}{*}{\textbf{Methods}} & \multirow{2}{*}{\textbf{\shortstack{Pretrain \\ Dataset \\ (Hours)}}} & \multicolumn{3}{c|}{\textbf{FACED}} & \multicolumn{3}{c|}{\textbf{Physionet-MI}} & \multicolumn{3}{c}{\textbf{SHU-MI}} \\
\cmidrule(lr){3-5} \cmidrule(lr){6-8} \cmidrule(l){9-11}
& & B. Acc. & AUC-PR & AUROC & B. Acc. & AUC-PR & AUROC & B. Acc. & Kappa & F1-W  \\
\midrule \midrule
BIOT {\scriptsize \cite{yang2023biot}} & 122 & 51.18 & 44.76 & 51.36 & 61.53 & 48.75 & 61.58 & 61.79 & 67.70 & 66.09 \\
LaBraM {\scriptsize \cite{jiang2024large}} & 2500 & 52.73 & 46.98 & 52.88 & 61.73 & 49.12 & 61.77 & 61.66 & 67.61 & 66.04 \\
\midrule
CBraMod {\scriptsize 25\% \cite{wang2025cbramod}} & 2250 & 55.34 & 49.54 & 55.63 & 61.83 & 49.12 & 61.91 & 62.75 & 68.84 & 68.86 \\
CBraMod {\scriptsize 100\% \cite{wang2025cbramod}} & 9000 & 55.09 & 50.41 & 56.18 & 64.17 & 52.22 & 64.27 & 63.70 & 71.39 & 69.88 \\
CBraMod-MTDP {\scriptsize 25\%} (Ours) & 2250 & 56.57 & 50.74 & 56.51 & 62.74 & 50.31 & 62.68 & \textbf{63.78} & \textbf{71.58} & \textbf{70.52} \\
CBraMod-MTDP {\scriptsize 100\%} (Ours) & 9000 & \textbf{56.95} & \textbf{51.29} & \textbf{57.35} & \textbf{64.57} & \textbf{52.75} & \textbf{64.70} & 62.84 & 71.11 & 69.07 \\ \noalign{\vspace{-4pt}} 
\scriptsize \textit{$\Delta =$ Ours 100\% - CBraMod 100\%} &  & {\color{gray} \scriptsize +1.86} & {\color{gray} \scriptsize +0.88} & {\color{gray} \scriptsize +1.17} & {\color{gray} \scriptsize +0.40} & {\color{gray} \scriptsize +0.53} & {\color{gray} \scriptsize +0.43} & {\color{gray} \scriptsize -0.86} & {\color{gray} \scriptsize -0.28} & {\color{gray} \scriptsize -0.81} \\[-0.4ex]
\bottomrule[1.5pt] \addlinespace[5pt]
\toprule[1.5pt]
\multirow{2}{*}{\textbf{Methods}} & \multirow{2}{*}{\textbf{\shortstack{Pretrain \\ Dataset \\ (Hours)}}} & \multicolumn{3}{c|}{\textbf{BCIC2020-3}} & \multicolumn{3}{c|}{\textbf{Mumtaz2016}} & \multicolumn{3}{c}{\textbf{MentalArithmetic}} \\
\cmidrule(lr){3-5} \cmidrule(lr){6-8} \cmidrule(l){9-11}
& & B. Acc. & AUC-PR & AUROC & B. Acc. & Kappa & F1-W & B. Acc. & Kappa & F1-W  \\
\midrule \midrule
BIOT {\scriptsize \cite{yang2023biot}} & 122 & 49.20 & 36.50 & 49.17 & 93.58 & 97.36 & 97.58 & 68.75 & 60.04 & 75.36 \\
LaBraM {\scriptsize \cite{jiang2024large}} & 2500 & 50.60 & 38.00 & 50.54 & 94.09 & 97.98 & 97.82 & 69.09 & 59.99 & 77.21 \\
\midrule
CBraMod {\scriptsize 25\% \cite{wang2025cbramod}} & 2250 & 44.00 & 30.00 & 44.03 & 90.31 & 97.87 & 97.56 & 64.24 & 46.91 & 74.94 \\
CBraMod {\scriptsize 100\% \cite{wang2025cbramod}} & 9000 & 53.73 & 42.16 & 53.83 & 95.60 & 99.23 & 99.21 & 72.56 & 62.67 & 79.05 \\
CBraMod-MTDP {\scriptsize 25\%} (Ours) & 2250 & 59.87 & 49.83 & 59.91 & 92.72 & 98.52 & 98.45 & 74.65 & 63.08 & 82.49 \\
CBraMod-MTDP {\scriptsize 100\%} (Ours) & 9000 & \textbf{62.53} & \textbf{53.17} & \textbf{62.57} & \textbf{95.85} & \textbf{99.55} & \textbf{99.55} & \textbf{77.43} & \textbf{73.30} & \textbf{87.37} \\ \noalign{\vspace{-4pt}} 
\scriptsize \textit{$\Delta =$ Ours 100\% - CBraMod 100\%} &  & {\color{gray} \scriptsize +8.80} & {\color{gray} \scriptsize +11.01} & {\color{gray} \scriptsize +8.74} & {\color{gray} \scriptsize +0.25} & {\color{gray} \scriptsize +0.32} & {\color{gray} \scriptsize +0.34} & {\color{gray} \scriptsize +4.87} & {\color{gray} \scriptsize +10.63} & {\color{gray} \scriptsize +8.32} \\[-0.4ex]
\bottomrule[1.5pt] \addlinespace[5pt]
\toprule[1.5pt]
\multirow{2}{*}{\textbf{Methods}} & \multirow{2}{*}{\textbf{\shortstack{Pretrain \\ Dataset \\ (Hours)}}} & \multicolumn{3}{c|}{\textbf{ISRUC}} & \multicolumn{3}{c|}{\textbf{TUEV}} & \multicolumn{3}{c}{\textbf{CHB-MIT}} \\
\cmidrule(lr){3-5} \cmidrule(lr){6-8} \cmidrule(l){9-11}
& & B. Acc. & AUC-PR & AUROC & B. Acc. & AUC-PR & AUROC & B. Acc. & Kappa & F1-W  \\
\midrule \midrule
BIOT {\scriptsize \cite{yang2023biot}} & 122 & 75.27 & 71.92 & 77.90 & 52.81 & 52.73 & 74.92 & 70.68 & 32.77 & 87.61 \\
LaBraM {\scriptsize \cite{jiang2024large}} & 2500 & 76.33 & 72.31 & 78.10 & 64.09 & 66.37 & 83.12 & 70.75 & 32.87 & 86.79 \\
\midrule
CBraMod {\scriptsize 25\% \cite{wang2025cbramod}} & 2250 & 77.19 & 73.97 & 79.58 & 64.21 & 64.09 & 81.25 & 72.52 & 40.20 & 89.31 \\
CBraMod {\scriptsize 100\% \cite{wang2025cbramod}} & 9000 & 78.65 & 74.42 & 80.11 & \textbf{66.59} & 67.44 & 83.31 & 73.98 & 36.89 & 88.92 \\
CBraMod-MTDP {\scriptsize 25\%} (Ours) & 2250 & \textbf{79.41} & \textbf{75.54} & \textbf{80.87} & 66.45 & 70.07 & 84.38 & \textbf{80.13} & 62.27 & \textbf{94.96} \\
CBraMod-MTDP {\scriptsize 100\%} (Ours) & 9000 & 79.38 & 75.12 & 80.49 & 64.88 & \textbf{71.03} & \textbf{84.65} & 75.42 & \textbf{65.84} & 94.21 \\ \noalign{\vspace{-4pt}} 
\scriptsize \textit{$\Delta =$ Ours 100\% - CBraMod 100\%} &  & {\color{gray} \scriptsize +0.73} & {\color{gray} \scriptsize +0.70} & {\color{gray} \scriptsize +0.38} & {\color{gray} \scriptsize -1.71} & {\color{gray} \scriptsize +3.59} & {\color{gray} \scriptsize +1.34} & {\color{gray} \scriptsize +1.44} & {\color{gray} \scriptsize +28.95} & {\color{gray} \scriptsize +5.29} \\[-0.4ex]
\bottomrule[1.5pt] \addlinespace[5pt]
\toprule[1.5pt]
\multirow{2}{*}{\textbf{Methods}} & \multirow{2}{*}{\textbf{\shortstack{Pretrain \\ Dataset \\ (Hours)}}} & \multicolumn{3}{c|}{\textbf{TUAB}} & \multicolumn{3}{c|}{\textbf{SEED-V}} & \multicolumn{3}{c}{\textbf{BCIC-IV-2a}} \\
\cmidrule(lr){3-5} \cmidrule(lr){6-8} \cmidrule(l){9-11}
& & B. Acc. & Kappa & F1-W & B. Acc. & AUC-PR & AUROC & B. Acc. & AUC-PR & AUROC  \\
\midrule \midrule
BIOT {\scriptsize \cite{yang2023biot}} & 122 & 79.59 & 87.92 & 88.15 & 38.37 & 22.61 & 38.56 & 47.48 & 29.97 & 46.07 \\
LaBraM {\scriptsize \cite{jiang2024large}} & 2500 & 81.40 & 89.65 & 90.22 & 39.76 & 23.86 & 39.74 & 48.69 & 31.59 & 47.58 \\
\midrule
CBraMod {\scriptsize 25\% \cite{wang2025cbramod}} & 2250 & 78.53 & 87.38 & 86.87 & 39.39 & 24.61 & 40.07 & 49.31 & 32.41 & 48.74 \\
CBraMod {\scriptsize 100\% \cite{wang2025cbramod}} & 9000 & \textbf{82.49} & \textbf{92.21} & \textbf{91.56} & 40.91 & 25.69 & 41.01 & 51.38 & 35.18 & 49.84 \\
CBraMod-MTDP {\scriptsize 25\%} (Ours) & 2250 & 79.81 & 88.30 & 87.65 & 41.33 & 27.17 & 42.14 & 59.63 & 46.18 & 58.81 \\
CBraMod-MTDP {\scriptsize 100\%} (Ours) & 9000 & 81.02 & 88.98 & 88.06 & \textbf{41.94} & \textbf{27.82} & \textbf{42.69} & \textbf{59.81} & \textbf{46.41} & \textbf{59.11} \\ \noalign{\vspace{-4pt}} 
\scriptsize \textit{$\Delta =$ Ours 100\% - CBraMod 100\%} &  & {\color{gray} \scriptsize -1.47} & {\color{gray} \scriptsize -3.23} & {\color{gray} \scriptsize -3.50} & {\color{gray} \scriptsize +1.03} & {\color{gray} \scriptsize +2.13} & {\color{gray} \scriptsize +1.68} & {\color{gray} \scriptsize +8.43} & {\color{gray} \scriptsize +11.23} & {\color{gray} \scriptsize +9.27} \\[-0.4ex]
\bottomrule[1.5pt] \addlinespace[5pt]
\end{tabular}
\end{table*}

\subsection{Experimental Setups}

\subsubsection{Teacher and Student Models}
For the student, we adopt CBraMod \cite{wang2025cbramod} as the EEG FM with 4M parameters, consisting of patch encoders, positional encoder, and criss-cross transformer. The architecture is detailed in \ref{tab:app_pretraininghyperparameters}. For teachers, we adopt DINOv3-ViT-B/16 \cite{simeoni2025dinov3} and Chronos-Bolt-Base \cite{ansari2024chronos}. DINOv3-ViT-B/16 is distilled from DINOv3-7B that is pretrained on 1.7 billion images. It is based on the ViT-B/16 \cite{dosovitskiy2020image} architecture with 87M parameters, consisting of a 12-layer transformer encoder. Chronos-Bolt-Base is a zero-shot time series forecasting model pretrained on nearly 100 billion time series observations. Chronos-Bolt-Base is based on the T5-Efficient-Base \cite{tay2021scale} architecture with 205M parameters, consisting of a 12-layer transformer encoder and 12-layer transformer decoder. We extract representations from the encoder only, as the decoder is designed for generative forecasting.
% DINOv3-ViT-B/16 is the distilled version of the original DINOv3-7B that consists of 7B parameters and is pretrained on 1.7 billion images. Specifically, 

\subsubsection{Preprocessing EEG for Teachers} \label{sec:eegpreprocessing}
While numerous strategies are available for preprocessing EEG signals for DINOv3 and Chronos, we adopt the Occam Razor's principle, prioritizing minimal computational overhead. 
DINOv3 expects images, which are represented as tensors of shape $\mathbb{R}^{3 \times H \times W }$, where $H$ and $W$ are the height and width. EEG signals are tensors of shape $\mathbb{R}^{C \times T}$, which can be considered as single-channel images with height $C$ and width $T$. We preprocesss EEG via min-max normalization to 0-1. We apply min-max normalization to scale the EEG signal to the $[0, 1]$ range, followed by a replication of the data over three channels. We also experimented with the spectrogram and continuous wavelet transform to provide additional phase-frequency information, but they introduced significant computational overhead and did not result in significant differences in performance. 
Chronos expects univariate time series, which are tensors of shape $\mathbb{R}^L$, where $L$ is the sequence length. The EEG channels are processed independently, and the mean representation across all channels is used.

\subsubsection{Pretraining and Downstream Datasets}

\textbf{Pretraining Dataset and Preprocessing.}
Following \cite{wang2025cbramod}, we adopt the Temple University Hospital EEG Corpus (TUEG) for pretraining \cite{obeid2016temple}, in order to ensure fair comparison between the self-supervised and the distillation pretraining strategy. Appendix Section \ref{sec:app_pretraindatasetpreprocessing} describes the preprocessing pipeline.

\textbf{Downstream Datasets.}
To assess the performance and generalizability of the proposed method, we conduct evaluations on 12 publicly available downstream datasets. We follow the preprocessing pipeline and subject-independent dataset splits adopted in CBraMod \cite{wang2025cbramod}. Table \ref{tab:downstreamdatasets} summarizes all the downstream datasets.

\subsubsection{Training and Evaluation Details}

\textbf{Pretraining Stage 1 Teacher Representation Fusion.}
The gating network consists of a 2-layer MLP with ReLU activation between the layers and softmax activation to normalize the weights. The prediction heads consist of a single fully connected layer. Mean Squared Error (MSE) loss is adopted as the masked latent denoising objective.

\textbf{Pretraining Stage 2 Knowledge Distillation.}
The pretrained weights for the gating network from stage 1 are loaded and frozen during stage 2 knowledge distillation. The projection head consists of a single fully connected layer. Cosine similarity is adopted as the distillation objective to minimize the distance between the projected student representations and the fused teacher representation.

\textbf{Training Setup.}
Across both stages, we adopted a batch size of 64 per GPU across four V100 GPUs, resulting in a global batch size of 256. To avoid redundant teacher inference during distillation, teacher representations were pre-computed and stored prior to the start of the distillation process. The gate is trained for one epoch in less than an hour. Distillation on a 25\% subset of the TUEG dataset over 40 epochs is completed in 21 hours. The loss curves for both are illustrated in Appendix Figure \ref{fig:app_loss_plots}. Hyperparameters for pretraining is detailed in Appendix Table \ref{tab:app_pretraininghyperparameters}. During distillation, the gating network assigns DINOv3 a weight of 0.56 and Chronos a weight of 0.43.

\textbf{Downstream Fine-tuning.}
Following CBraMod \cite{wang2025cbramod}, the last layer representations are flattened and passed to a classifier. The classifier consists of a 3-layer MLP with ELU activation and dropout. The model and classifier are jointly fine-tuned for each downstream dataset over 50 epochs. The best checkpoint is saved based on validation performance and evaluated on the test set.

\textbf{Baselines and Metrics.}
We select traditional EEG models and state-of-the-art EEG FMs as baselines. Please refer to Appendix Table \ref{tab:app_eegfmresults_part1}-\ref{tab:app_eegfmresults_part2} for all baselines. Our main baseline for comparison is CBraMod \cite{wang2025cbramod}, as we adopt an identical architecture and replace the self-supervised pretraining with the proposed MTDP pretraining. We report Balanced Accuracy, AUC-PR, and AUROC for binary classification and Balanced Accuracy, Cohen's Kappa, and Weighted-F1 for multi-class classification. 

\subsection{Experimental Results}

\subsubsection{EEG Foundation Model Comparison}
Experimental results on 12 downstream datasets are presented in Table \ref{tab:eegfmresults}. We present two versions of CBraMod pretrained on 25\% and 100\% of TUEG using the proposed MTDP framework. ``CBraMod-MTDP 25\%'' outperforms CBraMod on 9 out of 12 downstream datasets. When scaling the pretraining dataset to 100\%, we observe improved performance across all datasets, except SHU-MI and ISRUC. ``CBraMod-MTDP 100\%'' achieves better performance than CBraMod on 10 out of 12 downstream datasets. Compared to CBraMod, significant performance gains are observed for BCIC2020-3 (8.80/11.01/8.74\%), MentalArithmetic (4.87/10.63/8.32\%), CHB-MIT (1.44/28.95/5.29\%), and BCIC-IV-2a (8.43/11.23/9.27\%). These performance gains are notable, as we adopt an identical pretraining dataset, fine-tuning algorithm, and model architecture as CBraMod. The result highlights the merit of MTDP as a promising pretraining framework for EEG FMs.

\subsubsection{Teacher Representation Fusion Ablation}

\begin{table}[t]
    \centering
    \caption{Ablation of different distillation methods}
    \label{tab:distillationablationresults}
    \renewcommand{\arraystretch}{1.1}
    \begin{tabular}{l|ccc|c}
        \toprule[1.5pt]
        \textbf{Component} & \textbf{Set 1} & \textbf{Set 2} & \textbf{Set 3} & \textbf{Set 4} \\
        \midrule \midrule
        Chronos   & $\checkmark$ & $\times$     & $\checkmark$ & $\checkmark$ \\
        DINOv3      & $\times$     & $\checkmark$ & $\checkmark$ & $\checkmark$ \\
        Two-Stage & $\times$     & $\times$     & $\times$     & $\checkmark$ \\
        \midrule
        FACED     & 51.13          & 54.40          & 56.39          & \textbf{56.57}         \\
        PhysioNetMI&62.57          & 62.62          & 62.56          & \textbf{62.74}
             \\
        SHU-MI    & 62.67          & 62.27          & 62.11          & \textbf{63.78}         \\
        BCI2020-3 & 37.73          & 56.93          & 42.80          & \textbf{59.87}         \\
        \bottomrule[1.5pt]
    \end{tabular}
    \vspace{-10pt}
\end{table}

\begin{figure}[t]
    \centering
    \includegraphics[width=0.95\linewidth]{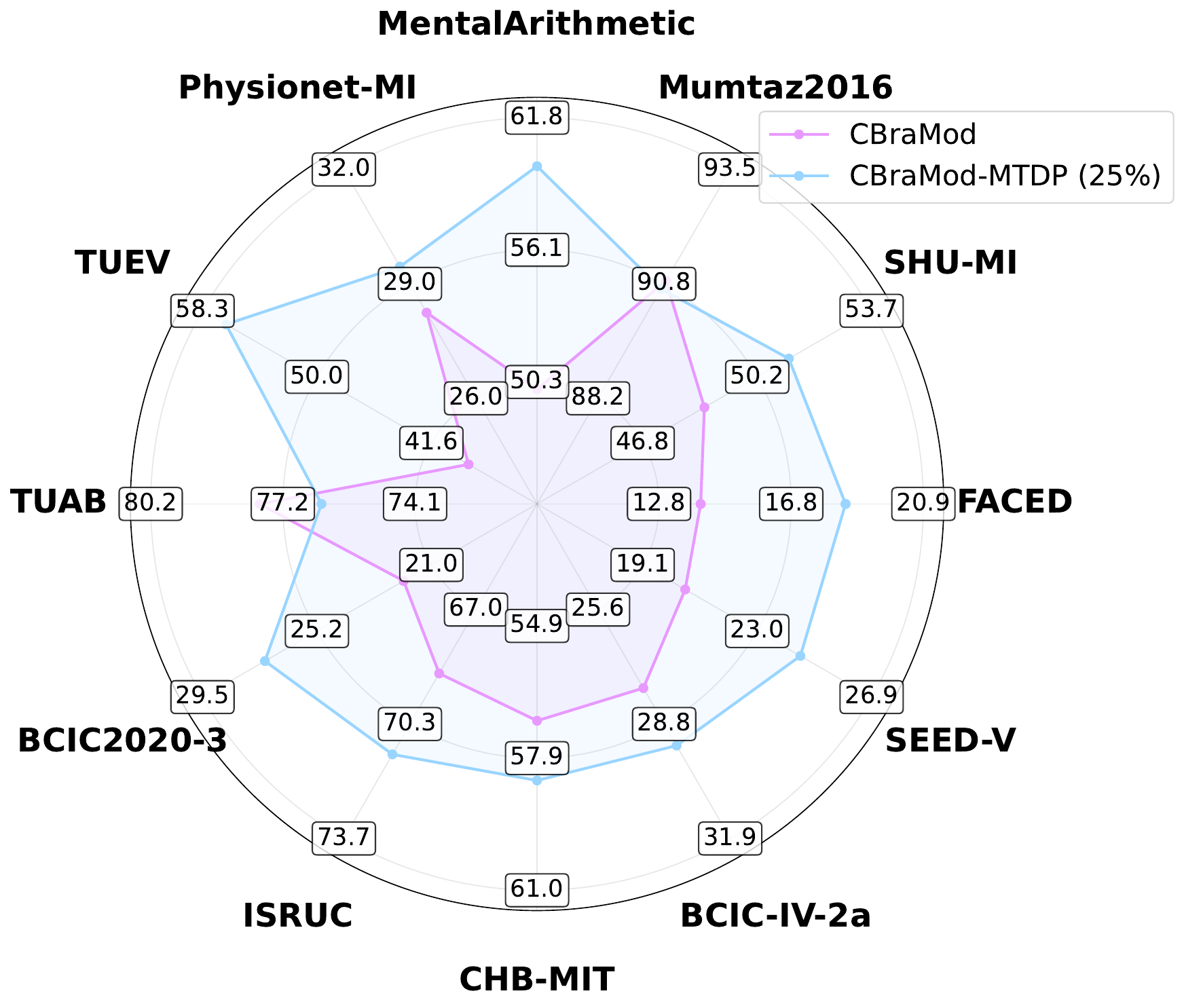}
    \caption{Linear probing performance of CBraMod and CBraMod-MTDP on EEG downstream tasks. Balanced Accuracy (\%).}
    \label{fig:linearprobingcbramodcbramodmtdp}
    \vspace{-12pt}
\end{figure}

In Table~\ref{tab:distillationablationresults}, we perform ablation studies to show the effectiveness of teacher fusion from stage 1 of the pretraining presented in Section \ref{sec:stage1pretrain}. All experiments are conducted with 25\% pretraining dataset. We remove teacher fusion and distill with one teacher only. Set 1 distills from Chronos. Set 2 distills from DINO. Similarly to AM-RADIO \cite{ranzinger2024radio}, we also distill from both teachers by summing the loss from each teacher, as denoted by set 3. Set 4 is the full MTDP framework proposed in Section \ref{sec:twostagedistillation}. 

Comparing set 1 and set 2, we observe that distilling from Chronos is better than distilling from DINOv3 on SHU-MI, but worse on FACED, PhysioNetMI and BCI2020-3. This suggests that the individual teachers provide representations that excel for different downstream tasks. In addition, we observe that set 3 performs better than set 1 and set 2 in FACED, worse performance in SHU-MI and PhysioNetMI, and between the two in BCI2020-3. This suggests that simply summing the losses from both teachers does not always yield better performance than distilling from a single teacher. Overall, set 4 achieves the best performance across the board. This highlights the effectiveness of the proposed framework in fusing teacher representations.

\subsubsection{Linear Probing Comparison}

Similar to Section \ref{sec:preliminaryexperiments}, we compare the representation quality of the reconstruction-pretrained CBraMod with the distillation-pretrained CBraMod-MTDP. We extract representations from frozen CBraMod and CBraMod-MTDP 25\%, then analyze the linear separability of these representations by training a logistic regression classifier. Figure \ref{fig:linearprobingcbramodcbramodmtdp} summarizes the linear probing performance. It is observed that CBraMod-MTDP 25\% outperforms CBraMod in 10 of 12 downstream datasets. These findings suggest that the proposed framework significantly enhances the model's ability to learn generalizable representations during pretraining.

\section{Conclusion}
In this paper, we present a new paradigm for EEG FM pretraining by distilling from mainstream FMs. The proposed MTDP framework demonstrates consistent improvement over the masked reconstruction pretraining counterpart on the vast majority of the downstream tasks. We hope that this work inspires the community to rethink the pretraining EEG FMs and embrace the integration of cross-domain insights.

\clearpage
\section*{Impact Statement}

This paper presents work whose goal is to advance the field of Machine
Learning. There are many potential societal consequences of our work, none
of which we feel must be specifically highlighted here.

\bibliography{example_paper}

\begin{thebibliography}{49}
\providecommand{\natexlab}[1]{#1}
\providecommand{\url}[1]{\texttt{#1}}
\expandafter\ifx\csname urlstyle\endcsname\relax
  \providecommand{\doi}[1]{doi: #1}\else
  \providecommand{\doi}{doi: \begingroup \urlstyle{rm}\Url}\fi

\bibitem[Aboalayon et~al.(2016)Aboalayon, Faezipour, Almuhammadi, and Moslehpour]{aboalayon2016sleep}
Aboalayon, K. A.~I., Faezipour, M., Almuhammadi, W.~S., and Moslehpour, S.
\newblock Sleep stage classification using eeg signal analysis: a comprehensive survey and new investigation.
\newblock \emph{Entropy}, 18\penalty0 (9):\penalty0 272, 2016.

\bibitem[{Al-Maamari} et~al.(2025){Al-Maamari}, Ben~Amor, Mitrovi{\'c}, and Granitzer]{al2025mixture}
{Al-Maamari}, M., Ben~Amor, M., Mitrovi{\'c}, J., and Granitzer, M.
\newblock Mixture of {{Modular Experts}}: {{Distilling Knowledge}} from a {{Multilingual Teacher}} into {{Specialized Modular Language Models}}.
\newblock In \emph{Proceedings of the 40th {{ACM}}/{{SIGAPP Symposium}} on {{Applied Computing}}}, pp.\  945--852. 2025.

\bibitem[Alotaiby et~al.(2014)Alotaiby, Alshebeili, Alshawi, Ahmad, and Abd El-Samie]{alotaiby2014eeg}
Alotaiby, T.~N., Alshebeili, S.~A., Alshawi, T., Ahmad, I., and Abd El-Samie, F.~E.
\newblock Eeg seizure detection and prediction algorithms: a survey.
\newblock \emph{EURASIP Journal on Advances in Signal Processing}, 2014\penalty0 (1):\penalty0 183, 2014.

\bibitem[Altaheri et~al.(2023)Altaheri, Muhammad, Alsulaiman, Amin, Altuwaijri, Abdul, Bencherif, and Faisal]{altaheri2023deep}
Altaheri, H., Muhammad, G., Alsulaiman, M., Amin, S.~U., Altuwaijri, G.~A., Abdul, W., Bencherif, M.~A., and Faisal, M.
\newblock Deep learning techniques for classification of electroencephalogram (eeg) motor imagery (mi) signals: A review.
\newblock \emph{Neural Computing and Applications}, 35\penalty0 (20):\penalty0 14681--14722, 2023.

\bibitem[Ansari et~al.(2024)Ansari, Stella, Turkmen, Zhang, Mercado, Shen, Shchur, Rangapuram, Arango, Kapoor, et~al.]{ansari2024chronos}
Ansari, A.~F., Stella, L., Turkmen, C., Zhang, X., Mercado, P., Shen, H., Shchur, O., Rangapuram, S.~S., Arango, S.~P., Kapoor, S., et~al.
\newblock Chronos: Learning the language of time series.
\newblock \emph{arXiv preprint arXiv:2403.07815}, 2024.

\bibitem[Berry et~al.(2017)Berry, Brooks, Gamaldo, Harding, Lloyd, Quan, Troester, and Vaughn]{berry2017aasm}
Berry, R.~B., Brooks, R., Gamaldo, C., Harding, S.~M., Lloyd, R.~M., Quan, S.~F., Troester, M.~T., and Vaughn, B.~V.
\newblock Aasm scoring manual updates for 2017 (version 2.4), 2017.

\bibitem[Chen et~al.(2024)Chen, Ren, Song, Wang, Wang, Li, and Qiu]{chen2024eegformer}
Chen, Y., Ren, K., Song, K., Wang, Y., Wang, Y., Li, D., and Qiu, L.
\newblock {{EEGFormer}}: {{Towards Transferable}} and {{Interpretable Large-Scale EEG Foundation Model}}, 2024.

\bibitem[Cui et~al.(2024)Cui, Jeong, Th{\"o}lke, Medani, Jerbi, Joshi, and Leahy]{cui2024neuro}
Cui, W., Jeong, W., Th{\"o}lke, P., Medani, T., Jerbi, K., Joshi, A.~A., and Leahy, R.~M.
\newblock Neuro-{{GPT}}: {{Towards A Foundation Model}} for {{EEG}}, 2024.

\bibitem[D{\"o}ner et~al.(2025)D{\"o}ner, Ingolfsson, Benini, and Li]{doner2025luna}
D{\"o}ner, B., Ingolfsson, T.~M., Benini, L., and Li, Y.
\newblock {{LUNA}}: {{Efficient}} and {{Topology-Agnostic Foundation Model}} for {{EEG Signal Analysis}}, 2025.

\bibitem[Dosovitskiy(2020)]{dosovitskiy2020image}
Dosovitskiy, A.
\newblock An image is worth 16x16 words: Transformers for image recognition at scale.
\newblock \emph{arXiv preprint arXiv:2010.11929}, 2020.

\bibitem[Fang et~al.(2025)Fang, Li, Zhou, Yu, Du, Qasem, Lu, Li, Zhang, and Goh]{fang2025neuript}
Fang, Z., Li, C., Zhou, H., Yu, S., Du, G., Qasem, A., Lu, Y., Li, J., Zhang, J., and Goh, S.~K.
\newblock {{NeurIPT}}: {{Foundation Model}} for {{Neural Interfaces}}, 2025.

\bibitem[Fisher et~al.(2017)Fisher, Cross, French, Higurashi, Hirsch, Jansen, Lagae, Mosh{\'e}, Peltola, Roulet~Perez, et~al.]{fisher2017operational}
Fisher, R.~S., Cross, J.~H., French, J.~A., Higurashi, N., Hirsch, E., Jansen, F.~E., Lagae, L., Mosh{\'e}, S.~L., Peltola, J., Roulet~Perez, E., et~al.
\newblock Operational classification of seizure types by the international league against epilepsy: Position paper of the ilae commission for classification and terminology.
\newblock \emph{Epilepsia}, 58\penalty0 (4):\penalty0 522--530, 2017.

\bibitem[Grattafiori et~al.(2024)Grattafiori, Dubey, Jauhri, Pandey, Kadian, Al-Dahle, Letman, Mathur, Schelten, Vaughan, et~al.]{grattafiori2024llama}
Grattafiori, A., Dubey, A., Jauhri, A., Pandey, A., Kadian, A., Al-Dahle, A., Letman, A., Mathur, A., Schelten, A., Vaughan, A., et~al.
\newblock The llama 3 herd of models.
\newblock \emph{arXiv preprint arXiv:2407.21783}, 2024.

\bibitem[Hinton et~al.(2015)Hinton, Vinyals, and Dean]{hinton2015distilling}
Hinton, G., Vinyals, O., and Dean, J.
\newblock Distilling the {{Knowledge}} in a {{Neural Network}}, 2015.

\bibitem[Huang et~al.(2024)Huang, Wang, He, Zitnik, and Zhang]{huang2024repurposing}
Huang, N., Wang, H., He, Z., Zitnik, M., and Zhang, X.
\newblock Repurposing {{Foundation Model}} for {{Generalizable Medical Time Series Classification}}, 2024.

\bibitem[Jiang et~al.(2024)Jiang, Zhao, and Lu]{jiang2024large}
Jiang, W.-B., Zhao, L.-M., and Lu, B.-L.
\newblock Large {{Brain Model}} for {{Learning Generic Representations}} with {{Tremendous EEG Data}} in {{BCI}}, 2024.

\bibitem[Jin et~al.(2024)Jin, Wang, Ma, Chu, Zhang, Shi, Chen, Liang, Li, Pan, and Wen]{jin2024timea}
Jin, M., Wang, S., Ma, L., Chu, Z., Zhang, J.~Y., Shi, X., Chen, P.-Y., Liang, Y., Li, Y.-F., Pan, S., and Wen, Q.
\newblock Time-{{LLM}}: {{Time Series Forecasting}} by {{Reprogramming Large Language Models}}, 2024.

\bibitem[Jing et~al.(2023)Jing, Ge, Hong, Fernandes, Lin, Yang, An, Struck, Herlopian, Karakis, et~al.]{jing2023development}
Jing, J., Ge, W., Hong, S., Fernandes, M.~B., Lin, Z., Yang, C., An, S., Struck, A.~F., Herlopian, A., Karakis, I., et~al.
\newblock Development of expert-level classification of seizures and rhythmic and periodic patterns during eeg interpretation.
\newblock \emph{Neurology}, 100\penalty0 (17):\penalty0 e1750--e1762, 2023.

\bibitem[Kim et~al.(2025)Kim, Chu, and Yang]{kim2025every}
Kim, G., Chu, G., and Yang, E.
\newblock Every {{Expert Matters}}: {{Towards Effective Knowledge Distillation}} for {{Mixture-of-Experts Language Models}}, 2025.

\bibitem[Lawhern et~al.(2018)Lawhern, Solon, Waytowich, Gordon, Hung, and Lance]{lawhern2018eegnet}
Lawhern, V.~J., Solon, A.~J., Waytowich, N.~R., Gordon, S.~M., Hung, C.~P., and Lance, B.~J.
\newblock Eegnet: a compact convolutional neural network for eeg-based brain--computer interfaces.
\newblock \emph{Journal of neural engineering}, 15\penalty0 (5):\penalty0 056013, 2018.

\bibitem[Li et~al.(2022{\natexlab{a}})Li, Ding, Zhang, and Xiu]{li2022motor}
Li, H., Ding, M., Zhang, R., and Xiu, C.
\newblock Motor imagery eeg classification algorithm based on cnn-lstm feature fusion network.
\newblock \emph{Biomedical signal processing and control}, 72:\penalty0 103342, 2022{\natexlab{a}}.

\bibitem[Li et~al.(2024{\natexlab{a}})Li, Liu, Cheng, Arcucci, and Hong]{li2024frozen}
Li, J., Liu, C., Cheng, S., Arcucci, R., and Hong, S.
\newblock Frozen {{Language Model Helps ECG Zero-Shot Learning}}.
\newblock In \emph{Medical {{Imaging}} with {{Deep Learning}}}, pp.\  402--415, 2024{\natexlab{a}}.

\bibitem[Li et~al.(2024{\natexlab{b}})Li, Zang, Wang, Dong, and Yang]{li2024multi}
Li, L., Zang, H., Wang, Y., Dong, Y., and Yang, L.
\newblock Multi-expert collaboration: {{Enhancing}} heterogeneous knowledge independence and alignment in knowledge distillation.
\newblock 2024{\natexlab{b}}.

\bibitem[Li et~al.(2022{\natexlab{b}})Li, Zhang, Tiwari, Song, Hu, Yang, Zhao, Kumar, and Marttinen]{li2022eeg}
Li, X., Zhang, Y., Tiwari, P., Song, D., Hu, B., Yang, M., Zhao, Z., Kumar, N., and Marttinen, P.
\newblock Eeg based emotion recognition: A tutorial and review.
\newblock \emph{ACM Computing Surveys}, 55\penalty0 (4):\penalty0 1--57, 2022{\natexlab{b}}.

\bibitem[Li et~al.(2023)Li, Li, and Yan]{li2023time}
Li, Z., Li, S., and Yan, X.
\newblock Time {{Series}} as {{Images}}: {{Vision Transformer}} for {{Irregularly Sampled Time Series}}, 2023.

\bibitem[Liu et~al.(2026)Liu, Chen, Chen, Cui, Wen, An, Luo, and Wu]{liu2026eegfoundationmodelsprogresses}
Liu, D., Chen, Y., Chen, Z., Cui, Z., Wen, Y., An, J., Luo, J., and Wu, D.
\newblock Eeg foundation models: Progresses, benchmarking, and open problems, 2026.
\newblock URL \url{https://arxiv.org/abs/2601.17883}.

\bibitem[Liu et~al.(2020)Liu, Zhang, and Wang]{liu2020adaptive}
Liu, Y., Zhang, W., and Wang, J.
\newblock Adaptive {{Multi-Teacher Multi-level Knowledge Distillation}}.
\newblock \emph{Neurocomputing}, 415:\penalty0 106--113, 2020.
\newblock \doi{10.1016/j.neucom.2020.07.048}.

\bibitem[Luo et~al.(2025)Luo, Chen, Salekin, and Rahman]{luo2025foundation}
Luo, Y., Chen, Y., Salekin, A., and Rahman, T.
\newblock Toward {{Foundation Model}} for {{Multivariate Wearable Sensing}} of {{Physiological Signals}}, 2025.

\bibitem[Obeid \& Picone(2016)Obeid and Picone]{obeid2016temple}
Obeid, I. and Picone, J.
\newblock The temple university hospital eeg data corpus.
\newblock \emph{Frontiers in neuroscience}, 10:\penalty0 196, 2016.

\bibitem[Ouahidi et~al.(2025)Ouahidi, Lys, Th{\"o}lke, Farrugia, Pasdeloup, Gripon, Jerbi, and Lioi]{ouahidi2025reve}
Ouahidi, Y.~E., Lys, J., Th{\"o}lke, P., Farrugia, N., Pasdeloup, B., Gripon, V., Jerbi, K., and Lioi, G.
\newblock {{REVE}}: {{A Foundation Model}} for {{EEG}} -- {{Adapting}} to {{Any Setup}} with {{Large-Scale Pretraining}} on 25,000 {{Subjects}}, 2025.

\bibitem[Peh et~al.(2022)Peh, Yao, and Dauwels]{peh2022transformer}
Peh, W.~Y., Yao, Y., and Dauwels, J.
\newblock Transformer convolutional neural networks for automated artifact detection in scalp eeg.
\newblock In \emph{2022 44th Annual International Conference of the IEEE Engineering in Medicine \& Biology Society (EMBC)}, pp.\  3599--3602. IEEE, 2022.

\bibitem[Pham et~al.(2022)Pham, Hoang, and Do]{pham2022collaborative}
Pham, C., Hoang, T., and Do, T.-T.
\newblock Collaborative {{Multi-Teacher Knowledge Distillation}} for {{Learning Low Bit-width Deep Neural Networks}}, 2022.

\bibitem[Ranzinger et~al.(2024)Ranzinger, Heinrich, Kautz, and Molchanov]{ranzinger2024radio}
Ranzinger, M., Heinrich, G., Kautz, J., and Molchanov, P.
\newblock Am-radio: Agglomerative vision foundation model reduce all domains into one.
\newblock In \emph{Proceedings of the IEEE/CVF conference on computer vision and pattern recognition}, pp.\  12490--12500, 2024.

\bibitem[Sim{\'e}oni et~al.(2025)Sim{\'e}oni, Vo, Seitzer, Baldassarre, Oquab, Jose, Khalidov, Szafraniec, Yi, Ramamonjisoa, Massa, Haziza, Wehrstedt, Wang, Darcet, Moutakanni, Sentana, Roberts, Vedaldi, Tolan, Brandt, Couprie, Mairal, J{\'e}gou, Labatut, and Bojanowski]{simeoni2025dinov3}
Sim{\'e}oni, O., Vo, H.~V., Seitzer, M., Baldassarre, F., Oquab, M., Jose, C., Khalidov, V., Szafraniec, M., Yi, S., Ramamonjisoa, M., Massa, F., Haziza, D., Wehrstedt, L., Wang, J., Darcet, T., Moutakanni, T., Sentana, L., Roberts, C., Vedaldi, A., Tolan, J., Brandt, J., Couprie, C., Mairal, J., J{\'e}gou, H., Labatut, P., and Bojanowski, P.
\newblock {{DINOv3}}, 2025.

\bibitem[Siuly et~al.(2017)Siuly, Li, and Zhang]{siuly2017electroencephalogram}
Siuly, S., Li, Y., and Zhang, Y.
\newblock Electroencephalogram (eeg) and its background.
\newblock In \emph{EEG signal analysis and classification: Techniques and applications}, pp.\  3--21. Springer, 2017.

\bibitem[Song et~al.(2021)Song, Jia, Yang, and Xie]{song2021transformer}
Song, Y., Jia, X., Yang, L., and Xie, L.
\newblock Transformer-based spatial-temporal feature learning for eeg decoding.
\newblock \emph{arXiv preprint arXiv:2106.11170}, 2021.

\bibitem[Song et~al.(2022)Song, Zheng, Liu, and Gao]{song2022eeg}
Song, Y., Zheng, Q., Liu, B., and Gao, X.
\newblock Eeg conformer: Convolutional transformer for eeg decoding and visualization.
\newblock \emph{IEEE Transactions on Neural Systems and Rehabilitation Engineering}, 31:\penalty0 710--719, 2022.

\bibitem[Tay et~al.(2021)Tay, Dehghani, Rao, Fedus, Abnar, Chung, Narang, Yogatama, Vaswani, and Metzler]{tay2021scale}
Tay, Y., Dehghani, M., Rao, J., Fedus, W., Abnar, S., Chung, H.~W., Narang, S., Yogatama, D., Vaswani, A., and Metzler, D.
\newblock Scale efficiently: Insights from pre-training and fine-tuning transformers.
\newblock \emph{arXiv preprint arXiv:2109.10686}, 2021.

\bibitem[Tegon et~al.(2025)Tegon, Ingolfsson, Wang, Benini, and Li]{tegon2025femba}
Tegon, A., Ingolfsson, T.~M., Wang, X., Benini, L., and Li, Y.
\newblock {{FEMBA}}: {{Efficient}} and {{Scalable EEG Analysis}} with a {{Bidirectional Mamba Foundation Model}}, 2025.

\bibitem[Wang et~al.(2025)Wang, Zhao, Luo, Zhou, Jiang, Li, Li, and Pan]{wang2025cbramod}
Wang, J., Zhao, S., Luo, Z., Zhou, Y., Jiang, H., Li, S., Li, T., and Pan, G.
\newblock {{CBraMod}}: {{A Criss-Cross Brain Foundation Model}} for {{EEG Decoding}}, 2025.

\bibitem[Xiao et~al.(2025)Xiao, Cui, Zhang, Chen, Wu, Thwaites, Woolgar, Zhou, and Zhang]{xiao2025brainomni}
Xiao, Q., Cui, Z., Zhang, C., Chen, S., Wu, W., Thwaites, A., Woolgar, A., Zhou, B., and Zhang, C.
\newblock {{BrainOmni}}: {{A Brain Foundation Model}} for {{Unified EEG}} and {{MEG Signals}}, 2025.

\bibitem[Xie et~al.(2024)Xie, Zhang, Zhuang, Shi, Liu, Gu, and Zhang]{xie2024mode}
Xie, Z., Zhang, Y., Zhuang, C., Shi, Q., Liu, Z., Gu, J., and Zhang, G.
\newblock {{MoDE}}: {{A Mixture-of-Experts Model}} with {{Mutual Distillation}} among the {{Experts}}, 2024.

\bibitem[Yang et~al.(2021)Yang, Xiao, Westover, and Sun]{yang2021self}
Yang, C., Xiao, D., Westover, M.~B., and Sun, J.
\newblock Self-supervised eeg representation learning for automatic sleep staging.
\newblock \emph{arXiv preprint arXiv:2110.15278}, 2021.

\bibitem[Yang et~al.(2023)Yang, Westover, and Sun]{yang2023biot}
Yang, C., Westover, M., and Sun, J.
\newblock {{BIOT}}: {{Biosignal Transformer}} for {{Cross-data Learning}} in the {{Wild}}.
\newblock \emph{Advances in Neural Information Processing Systems}, 36:\penalty0 78240--78260, 2023.

\bibitem[Yu \& Wang(2025)Yu and Wang]{yu2025healthcare}
Yu, T. and Wang, G.
\newblock Healthcare {{Event Prediction}} via {{Iterative}} in-{{Context Learning}} of {{LLM}}.
\newblock In \emph{2025 10th {{International Conference}} on {{Cloud Computing}} and {{Big Data Analytics}} ({{ICCCBDA}})}, pp.\  205--210, 2025.
\newblock \doi{10.1109/ICCCBDA64898.2025.11030505}.

\bibitem[Yuan et~al.(2024)Yuan, Zhang, Chen, Gu, and Yang]{yuan2024brant}
Yuan, Z., Zhang, D., Chen, J., Gu, G., and Yang, Y.
\newblock Brant-2: {{Foundation Model}} for {{Brain Signals}}, 2024.

\bibitem[Yuan et~al.(2025)Yuan, Shen, Li, Yu, Tan, and Yang]{yuan2025brainwave}
Yuan, Z., Shen, F., Li, M., Yu, Y., Tan, C., and Yang, Y.
\newblock {{BrainWave}}: {{A Brain Signal Foundation Model}} for {{Clinical Applications}}, 2025.

\bibitem[Zhang et~al.(2023)Zhang, Yuan, Yang, Chen, Wang, and Li]{zhang2023brant}
Zhang, D., Yuan, Z., Yang, Y., Chen, J., Wang, J., and Li, Y.
\newblock Brant: {{Foundation Model}} for {{Intracranial Neural Signal}}.
\newblock In \emph{Thirty-Seventh {{Conference}} on {{Neural Information Processing Systems}}}, 2023.

\bibitem[Zhong et~al.(2025)Zhong, Ruan, Jin, Li, Wen, and Liang]{zhong2025time}
Zhong, S., Ruan, W., Jin, M., Li, H., Wen, Q., and Liang, Y.
\newblock Time-{{VLM}}: {{Exploring Multimodal Vision-Language Models}} for {{Augmented Time Series Forecasting}}, 2025.

\end{thebibliography}
\bibliographystyle{icml2026}

%%%%%%%%%%%%%%%%%%%%%%%%%%%%%%%%%%%%%%%%%%%%%%%%%%%%%%%%%%%%%%%%%%%%%%%%%%%%%%%
%%%%%%%%%%%%%%%%%%%%%%%%%%%%%%%%%%%%%%%%%%%%%%%%%%%%%%%%%%%%%%%%%%%%%%%%%%%%%%%
% APPENDIX
%%%%%%%%%%%%%%%%%%%%%%%%%%%%%%%%%%%%%%%%%%%%%%%%%%%%%%%%%%%%%%%%%%%%%%%%%%%%%%%
%%%%%%%%%%%%%%%%%%%%%%%%%%%%%%%%%%%%%%%%%%%%%%%%%%%%%%%%%%%%%%%%%%%%%%%%%%%%%%%
\newpage
\appendix
\onecolumn
\section{Experiment Setting}

\subsection{Pretraining Dataset Preprocessing} \label{sec:app_pretraindatasetpreprocessing}
 We select 16 channels (Fp1, Fp2, F7, F3, F4, F8, T3, C3, C4, T4, T5, P3, P4, T6, O1, O2), apply band-pass (0.3-75 \unit{\hertz}) and notch (60 \unit{\hertz}) filter, resample to 200 \unit{\hertz}, segment into 30-second non-overlapping segments, discard samples with absolute amplitude exceeding 100 \unit{\micro\volt}, and normalize to -1 to 1 range by setting unit to 100 \unit{\micro\volt}. Unlike CBraMod, we apply an additional step to remove TUAB and TUEV samples to avoid downstream data contamination. This results in a total of 1,059,523 samples, which is around 8,829 hours in total.
 
\subsection{Pretraining Hyperparameters}
Table \ref{tab:app_pretraininghyperparameters} summarizes the hyperparameters used for pretraining.

\begin{table}[!h]
\centering
\caption{Pretraining Hyperparameters}
\label{tab:app_pretraininghyperparameters}
\small % Slightly smaller font to ensure the two columns fit comfortably
\begin{tabular}{@{}lc@{}}
\toprule[1.5pt]
\textbf{Hyperparameters} & \textbf{Settings} \\ \midrule

% --- DATA SECTION ---
\multicolumn{2}{c}{\textbf{Data}} \\ \midrule
Channels & 16 \\
Time points & 6000 \\
Patch dimension & 200 \\
Sequence length & 30 \unit{\second} \\
\midrule

% --- MODEL SECTION ---
\multicolumn{2}{c}{\textbf{CBraMod Model Architecture}} \\ \midrule
\textbf{Patch Encoder (Time-Domain)} & \\
\quad Input dimension & \{1, 25, 25\} \\
\quad Output dimension & \{25, 25, 25\} \\
\quad Kernel size & \{49, 3, 3\} \\
\quad Stride & \{25, 1, 1\} \\
\quad Padding & \{24, 1, 1\} \\
\textbf{Patch Encoder (Freq-Domain)} & \\
\quad FFT function & torch.fft.rfft \\
\quad Fully-connected layer & (101, 200) \\
\textbf{Positional Encoder} & \\
\quad Input dimension & 200 \\
\quad Output dimension & 200 \\
\quad Kernel size & (19, 7) \\
\quad Stride & (1, 1) \\
\quad Padding & (9, 3) \\
\textbf{Criss-cross Transformer} & \\
\quad Layers & 12 \\
\quad Hidden dimension & 200 \\
\quad Heads & 8 \\
\quad S-Attention heads & 4 \\
\quad T-Attention heads & 4 \\
\quad Feed-forward dimension & 800 \\ 
\bottomrule[1.5pt]
\end{tabular}
\hspace{1cm} % Adds spacing between the two tables
\begin{tabular}{@{}lc@{}}
\toprule[1.5pt]
\textbf{Hyperparameters} & \textbf{Settings} \\ \midrule

% --- GATE TRAINING SECTION ---
\multicolumn{2}{c}{\textbf{Stage 1 Teacher Representation Fusion}} \\ \midrule
Epochs & 1 \\
Batch size & 256 \\
Optimizer & AdamW \\
Learning rate & 5e-4 \\
Adam $\beta$ & (0.9, 0.999) \\
Adam $\epsilon$ & 1e-8 \\
Weight decay & 5e-2 \\
Clipping gradient norm & 1 \\
Weights init & Kaiming normalization \\ 
Loss & MSELoss \\
\midrule

% --- DISTILLATION SECTION ---
\multicolumn{2}{c}{\textbf{Stage 2 Knowledge Distillation}} \\ \midrule
Epochs & 40 \\
Batch size & 256 \\
Optimizer & AdamW \\
Learning rate & 5e-4 \\
Adam $\beta$ & (0.9, 0.999) \\
Adam $\epsilon$ & 1e-8 \\
Weight decay & 5e-2 \\
Scheduler & CosineAnnealingLR \\
Cosine cycle epochs & 40 \\
Minimal learning rate & 1e-5 \\
Clipping gradient norm & 1 \\
Weights init & Kaiming normalization \\ 
Loss & CosineEmbeddingLoss \\

\bottomrule[1.5pt]
\end{tabular}
\end{table}

\subsubsection{Training Loss Curves}
\begin{figure}[!h]
    \centering
    \includegraphics[width=\linewidth]{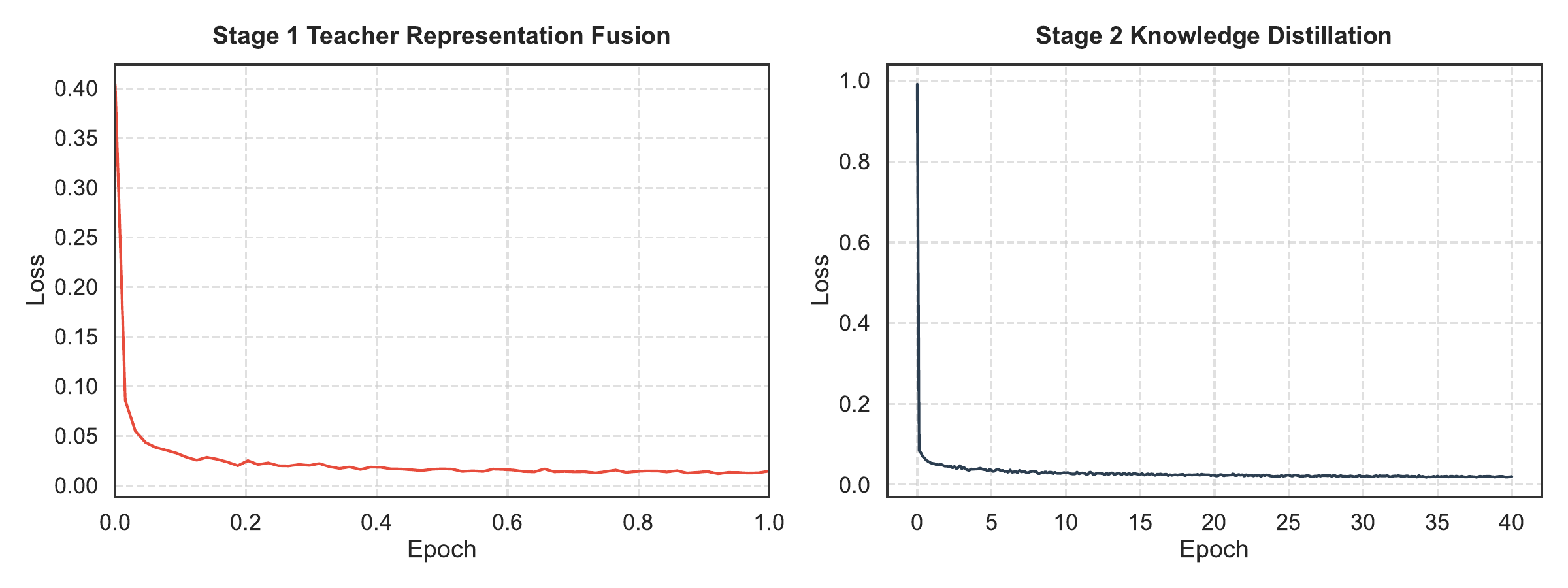}
    \caption{Loss curve of stage 1 and stage 2 pretraining}
    \label{fig:app_loss_plots}
\end{figure}

Figure \ref{fig:app_loss_plots} plots the loss curve of stage 1 and stage 2 pretraining.
\subsection{Fine-tuning Hyperparameters}
We adopt the fine-tuning code from CBraMod without any modification. We load our distilled checkpoint instead of the self-supervised checkpoint provided by CBraMod. Following CBraMod, we perform a grid search over three different learning rates and a binary multi-LR setting. Multi-LR adjusts the classifier learning rate to be five times larger. Table \ref{tab:app_finetuninghyperparameters} summarizes the fine-tuning hyperparameters.
\begin{table}[!h]
\centering
\caption{Fine-tuning Hyperparameters}
\label{tab:app_finetuninghyperparameters}
\begin{tabular}{lc}
\toprule
\textbf{Hyperparameters} & \textbf{Settings} \\ \midrule
Epochs & 50 \\
Batch size & 64 \\
Dropout & 0.1 \\
Optimizer & AdamW \\
Bakbone Learning rate & \{5e-5, 1e-4, 5e-4\} \\
Multi-LR & \{Yes, No\} \\
Adam $\beta$ & (0.9, 0.999) \\
Adam $\epsilon$ & 1e-8 \\
Weight decay & 5e-2 \\
Scheduler & CosineAnnealingLR \\
Cosine cycle epochs & 50 \\
Minimal learning rate & 1e-6 \\
Clipping gradient norm & 1 \\
Label smoothing (multi-class classification) & 0.1 \\ \bottomrule
\end{tabular}
\end{table}

\section{Comparison with Traditional EEG Models and EEG Foundation Models}
We compare against traditional EEG models and EEG FMs. The traditional models include: EEGNet \cite{lawhern2018eegnet}, EEGConformer \cite{song2022eeg}, SPaRCNet \cite{jing2023development}, ContraWR \cite{yang2021self}, CNN-Transformer \cite{peh2022transformer}, FFCL \cite{li2022motor}, ST-Transformer \cite{song2021transformer}. The EEG FMs include BIOT \cite{yang2023biot} and LaBraM \cite{jiang2024large}. Table \ref{tab:app_eegfmresults_part1}-\ref{tab:app_eegfmresults_part2} presents the full results.

\begin{table*}[!ht]
\centering
\caption{Comparison of EEG foundation model performance on 12 downstream tasks (Part 1/2). ``25\%'' and ``100\%'' indicate pretraining dataset size. Metrics acronyms: Balanced Accuracy (B. Acc.), Weighted F1 (F1-W), Cohen's Kappa (Kappa).}
\label{tab:app_eegfmresults_part1}
\small
\setlength{\tabcolsep}{3.5pt}
\renewcommand{\arraystretch}{1.2}
\begin{tabular}{l | c | ccc | ccc | ccc}
\toprule[1.5pt]
\multirow{2}{*}{\textbf{Methods}} & \multirow{2}{*}{\textbf{\shortstack{Pretrain \\ Dataset \\ (Hours)}}} & \multicolumn{3}{c|}{\textbf{FACED}} & \multicolumn{3}{c|}{\textbf{Physionet-MI}} & \multicolumn{3}{c}{\textbf{SHU-MI}} \\
\cmidrule(lr){3-5} \cmidrule(lr){6-8} \cmidrule(l){9-11}
& & B. Acc. & AUC-PR & AUROC & B. Acc. & AUC-PR & AUROC & B. Acc. & Kappa & F1-W  \\
\midrule \midrule
EEGNet & 0 & 40.90 & 33.42 & 41.24 & 58.14 & 44.68 & 57.96 & 58.89 & 63.11 & 62.83 \\
EEGConformer & 0 & 45.59 & 38.58 & 45.14 & 60.49 & 47.36 & 60.62 & 59.00 & 63.70 & 63.51 \\
SPaRCNet & 0 & 46.73 & 39.78 & 47.29 & 59.32 & 45.64 & 59.37 & 59.78 & 65.10 & 64.31 \\
ContraWR & 0 & 48.87 & 42.31 & 48.84 & 58.92 & 45.27 & 59.18 & 58.73 & 63.15 & 62.73 \\
CNN-Transformer & 0 & 46.97 & 40.17 & 47.20 & 60.53 & 47.25 & 60.41 & 59.75 & 64.12 & 63.23 \\
FFCL & 0 & 46.73 & 39.87 & 46.99 & 57.26 & 43.23 & 57.01 & 56.92 & 59.43 & 60.14 \\
ST-Transformer & 0 & 48.10 & 41.37 & 47.95 & 60.35 & 47.12 & 60.53 & 59.92 & 63.94 & 64.31 \\
\midrule
BIOT {\scriptsize \cite{yang2023biot}} & 122 & 51.18 & 44.76 & 51.36 & 61.53 & 48.75 & 61.58 & 61.79 & 67.70 & 66.09 \\
LaBraM {\scriptsize \cite{jiang2024large}} & 2500 & 52.73 & 46.98 & 52.88 & 61.73 & 49.12 & 61.77 & 61.66 & 67.61 & 66.04 \\
\midrule
CBraMod {\scriptsize 25\% \cite{wang2025cbramod}} & 2250 & 55.34 & 49.54 & 55.63 & 61.83 & 49.12 & 61.91 & 62.75 & 68.84 & 68.86 \\
CBraMod {\scriptsize 100\% \cite{wang2025cbramod}} & 9000 & 55.09 & 50.41 & 56.18 & 64.17 & 52.22 & 64.27 & 63.70 & 71.39 & 69.88 \\
CBraMod-MTDP {\scriptsize 25\%} (Ours) & 2250 & 56.57 & 50.74 & 56.51 & 62.74 & 50.31 & 62.68 & \textbf{63.78} & \textbf{71.58} & \textbf{70.52} \\
CBraMod-MTDP {\scriptsize 100\%} (Ours) & 9000 & \textbf{56.95} & \textbf{51.29} & \textbf{57.35} & \textbf{64.57} & \textbf{52.75} & \textbf{64.70} & 62.84 & 71.11 & 69.07 \\ \noalign{\vspace{-4pt}} 
\scriptsize \textit{$\Delta =$ Ours 100\% - CBraMod 100\%} &  & {\color{gray} \scriptsize +1.86} & {\color{gray} \scriptsize +0.88} & {\color{gray} \scriptsize +1.17} & {\color{gray} \scriptsize +0.40} & {\color{gray} \scriptsize +0.53} & {\color{gray} \scriptsize +0.43} & {\color{gray} \scriptsize -0.86} & {\color{gray} \scriptsize -0.28} & {\color{gray} \scriptsize -0.81} \\[-0.4ex]
\bottomrule[1.5pt] \addlinespace[5pt]
\toprule[1.5pt]
\multirow{2}{*}{\textbf{Methods}} & \multirow{2}{*}{\textbf{\shortstack{Pretrain \\ Dataset \\ (Hours)}}} & \multicolumn{3}{c|}{\textbf{BCIC2020-3}} & \multicolumn{3}{c|}{\textbf{Mumtaz2016}} & \multicolumn{3}{c}{\textbf{MentalArithmetic}} \\
\cmidrule(lr){3-5} \cmidrule(lr){6-8} \cmidrule(l){9-11}
& & B. Acc. & AUC-PR & AUROC & B. Acc. & Kappa & F1-W & B. Acc. & Kappa & F1-W  \\
\midrule \midrule
EEGNet & 0 & 44.13 & 30.16 & 44.13 & 92.32 & 96.26 & 96.39 & 67.70 & 57.63 & 73.21 \\
EEGConformer & 0 & 45.06 & 31.33 & 44.88 & 93.08 & 96.84 & 97.02 & 68.05 & 58.29 & 74.24 \\
SPaRCNet & 0 & 44.26 & 30.33 & 44.20 & 93.16 & 97.54 & 97.81 & 68.79 & 58.25 & 74.18 \\
ContraWR & 0 & 42.57 & 30.78 & 44.07 & 91.95 & 95.89 & 96.21 & 66.31 & 57.87 & 73.32 \\
CNN-Transformer & 0 & 45.33 & 31.66 & 45.06 & 93.05 & 97.57 & 97.42 & 67.79 & 57.77 & 72.58 \\
FFCL & 0 & 46.78 & 33.01 & 46.89 & 93.14 & 97.17 & 97.53 & 67.98 & 57.86 & 73.30 \\
ST-Transformer & 0 & 41.26 & 29.41 & 42.47 & 91.35 & 95.78 & 95.94 & 66.31 & 56.72 & 71.32 \\
\midrule
BIOT {\scriptsize \cite{yang2023biot}} & 122 & 49.20 & 36.50 & 49.17 & 93.58 & 97.36 & 97.58 & 68.75 & 60.04 & 75.36 \\
LaBraM {\scriptsize \cite{jiang2024large}} & 2500 & 50.60 & 38.00 & 50.54 & 94.09 & 97.98 & 97.82 & 69.09 & 59.99 & 77.21 \\
\midrule
CBraMod {\scriptsize 25\% \cite{wang2025cbramod}} & 2250 & 44.00 & 30.00 & 44.03 & 90.31 & 97.87 & 97.56 & 64.24 & 46.91 & 74.94 \\
CBraMod {\scriptsize 100\% \cite{wang2025cbramod}} & 9000 & 53.73 & 42.16 & 53.83 & 95.60 & 99.23 & 99.21 & 72.56 & 62.67 & 79.05 \\
CBraMod-MTDP {\scriptsize 25\%} (Ours) & 2250 & 59.87 & 49.83 & 59.91 & 92.72 & 98.52 & 98.45 & 74.65 & 63.08 & 82.49 \\
CBraMod-MTDP {\scriptsize 100\%} (Ours) & 9000 & \textbf{62.53} & \textbf{53.17} & \textbf{62.57} & \textbf{95.85} & \textbf{99.55} & \textbf{99.55} & \textbf{77.43} & \textbf{73.30} & \textbf{87.37} \\ \noalign{\vspace{-4pt}} 
\scriptsize \textit{$\Delta =$ Ours 100\% - CBraMod 100\%} &  & {\color{gray} \scriptsize +8.80} & {\color{gray} \scriptsize +11.01} & {\color{gray} \scriptsize +8.74} & {\color{gray} \scriptsize +0.25} & {\color{gray} \scriptsize +0.32} & {\color{gray} \scriptsize +0.34} & {\color{gray} \scriptsize +4.87} & {\color{gray} \scriptsize +10.63} & {\color{gray} \scriptsize +8.32} \\[-0.4ex]
\bottomrule[1.5pt] \addlinespace[5pt]
\end{tabular}
\end{table*}
\clearpage
\begin{table*}[!ht]
\centering
\caption{Comparison of EEG foundation model performance on 12 downstream tasks (Part 2/2). ``25\%'' and ``100\%'' indicate pretraining dataset size. Metrics acronyms: Balanced Accuracy (B. Acc.), Weighted F1 (F1-W), Cohen's Kappa (Kappa).}
\label{tab:app_eegfmresults_part2}
\small
\setlength{\tabcolsep}{3.5pt}
\renewcommand{\arraystretch}{1.2}
\begin{tabular}{l | c | ccc | ccc | ccc}
\toprule[1.5pt]
\multirow{2}{*}{\textbf{Methods}} & \multirow{2}{*}{\textbf{\shortstack{Pretrain \\ Dataset \\ (Hours)}}} & \multicolumn{3}{c|}{\textbf{ISRUC}} & \multicolumn{3}{c|}{\textbf{TUEV}} & \multicolumn{3}{c}{\textbf{CHB-MIT}} \\
\cmidrule(lr){3-5} \cmidrule(lr){6-8} \cmidrule(l){9-11}
& & B. Acc. & AUC-PR & AUROC & B. Acc. & AUC-PR & AUROC & B. Acc. & Kappa & F1-W  \\
\midrule \midrule
EEGNet & 0 & 71.54 & 70.40 & 75.13 & 38.76 & 35.77 & 65.39 & 56.58 & 19.14 & 80.48 \\
EEGConformer & 0 & 74.00 & 71.43 & 76.34 & 40.74 & 39.67 & 69.83 & 59.76 & 22.09 & 82.26 \\
SPaRCNet & 0 & 74.87 & 70.97 & 76.24 & 41.61 & 42.33 & 70.24 & 58.76 & 12.47 & 81.43 \\
ContraWR & 0 & 74.02 & 71.78 & 76.10 & 43.84 & 39.12 & 68.93 & 63.44 & 22.64 & 80.97 \\
CNN-Transformer & 0 & 73.63 & 71.29 & 77.19 & 40.87 & 38.15 & 68.54 & 63.89 & 24.79 & 86.62 \\
FFCL & 0 & 72.77 & 70.16 & 76.14 & 39.79 & 37.32 & 67.83 & 62.62 & 20.49 & 82.71 \\
ST-Transformer & 0 & 73.81 & 70.13 & 76.81 & 39.84 & 37.65 & 68.23 & 59.15 & 14.22 & 82.37 \\
\midrule
BIOT {\scriptsize \cite{yang2023biot}} & 122 & 75.27 & 71.92 & 77.90 & 52.81 & 52.73 & 74.92 & 70.68 & 32.77 & 87.61 \\
LaBraM {\scriptsize \cite{jiang2024large}} & 2500 & 76.33 & 72.31 & 78.10 & 64.09 & 66.37 & 83.12 & 70.75 & 32.87 & 86.79 \\
\midrule
CBraMod {\scriptsize 25\% \cite{wang2025cbramod}} & 2250 & 77.19 & 73.97 & 79.58 & 64.21 & 64.09 & 81.25 & 72.52 & 40.20 & 89.31 \\
CBraMod {\scriptsize 100\% \cite{wang2025cbramod}} & 9000 & 78.65 & 74.42 & 80.11 & \textbf{66.59} & 67.44 & 83.31 & 73.98 & 36.89 & 88.92 \\
CBraMod-MTDP {\scriptsize 25\%} (Ours) & 2250 & \textbf{79.41} & \textbf{75.54} & \textbf{80.87} & 66.45 & 70.07 & 84.38 & \textbf{80.13} & 62.27 & \textbf{94.96} \\
CBraMod-MTDP {\scriptsize 100\%} (Ours) & 9000 & 79.38 & 75.12 & 80.49 & 64.88 & \textbf{71.03} & \textbf{84.65} & 75.42 & \textbf{65.84} & 94.21 \\ \noalign{\vspace{-4pt}} 
\scriptsize \textit{$\Delta =$ Ours 100\% - CBraMod 100\%} &  & {\color{gray} \scriptsize +0.73} & {\color{gray} \scriptsize +0.70} & {\color{gray} \scriptsize +0.38} & {\color{gray} \scriptsize -1.71} & {\color{gray} \scriptsize +3.59} & {\color{gray} \scriptsize +1.34} & {\color{gray} \scriptsize +1.44} & {\color{gray} \scriptsize +28.95} & {\color{gray} \scriptsize +5.29} \\[-0.4ex]
\bottomrule[1.5pt] \addlinespace[5pt]
\toprule[1.5pt]
\multirow{2}{*}{\textbf{Methods}} & \multirow{2}{*}{\textbf{\shortstack{Pretrain \\ Dataset \\ (Hours)}}} & \multicolumn{3}{c|}{\textbf{TUAB}} & \multicolumn{3}{c|}{\textbf{SEED-V}} & \multicolumn{3}{c}{\textbf{BCIC-IV-2a}} \\
\cmidrule(lr){3-5} \cmidrule(lr){6-8} \cmidrule(l){9-11}
& & B. Acc. & Kappa & F1-W & B. Acc. & AUC-PR & AUROC & B. Acc. & AUC-PR & AUROC  \\
\midrule \midrule
EEGNet & 0 & 76.42 & 82.99 & 84.12 & 29.61 & 10.06 & 27.49 & 44.82 & 26.93 & 42.26 \\
EEGConformer & 0 & 77.58 & 84.27 & 84.45 & 35.37 & 17.72 & 34.87 & 46.96 & 29.24 & 45.33 \\
SPaRCNet & 0 & 78.96 & 84.14 & 86.76 & 29.49 & 11.21 & 29.79 & 46.35 & 28.47 & 44.32 \\
ContraWR & 0 & 77.46 & 84.21 & 84.56 & 35.46 & 19.05 & 35.44 & 46.78 & 29.05 & 44.13 \\
CNN-Transformer & 0 & 77.77 & 84.33 & 84.61 & 36.78 & 20.72 & 36.42 & 46.00 & 28.00 & 44.60 \\
FFCL & 0 & 78.48 & 84.48 & 85.69 & 36.41 & 20.78 & 36.45 & 44.70 & 26.27 & 42.38 \\
ST-Transformer & 0 & 79.66 & 85.21 & 87.07 & 30.52 & 10.83 & 28.33 & 45.75 & 27.33 & 44.71 \\
\midrule
BIOT {\scriptsize \cite{yang2023biot}} & 122 & 79.59 & 87.92 & 88.15 & 38.37 & 22.61 & 38.56 & 47.48 & 29.97 & 46.07 \\
LaBraM {\scriptsize \cite{jiang2024large}} & 2500 & 81.40 & 89.65 & 90.22 & 39.76 & 23.86 & 39.74 & 48.69 & 31.59 & 47.58 \\
\midrule
CBraMod {\scriptsize 25\% \cite{wang2025cbramod}} & 2250 & 78.53 & 87.38 & 86.87 & 39.39 & 24.61 & 40.07 & 49.31 & 32.41 & 48.74 \\
CBraMod {\scriptsize 100\% \cite{wang2025cbramod}} & 9000 & \textbf{82.49} & \textbf{92.21} & \textbf{91.56} & 40.91 & 25.69 & 41.01 & 51.38 & 35.18 & 49.84 \\
CBraMod-MTDP {\scriptsize 25\%} (Ours) & 2250 & 79.81 & 88.30 & 87.65 & 41.33 & 27.17 & 42.14 & 59.63 & 46.18 & 58.81 \\
CBraMod-MTDP {\scriptsize 100\%} (Ours) & 9000 & 81.02 & 88.98 & 88.06 & \textbf{41.94} & \textbf{27.82} & \textbf{42.69} & \textbf{59.81} & \textbf{46.41} & \textbf{59.11} \\ \noalign{\vspace{-4pt}} 
\scriptsize \textit{$\Delta =$ Ours 100\% - CBraMod 100\%} &  & {\color{gray} \scriptsize -1.47} & {\color{gray} \scriptsize -3.23} & {\color{gray} \scriptsize -3.50} & {\color{gray} \scriptsize +1.03} & {\color{gray} \scriptsize +2.13} & {\color{gray} \scriptsize +1.68} & {\color{gray} \scriptsize +8.43} & {\color{gray} \scriptsize +11.23} & {\color{gray} \scriptsize +9.27} \\[-0.4ex]
\bottomrule[1.5pt] \addlinespace[5pt]
\end{tabular}
\end{table*}

\section{Limitations}
The use of DINOv3 and Chronos incurs additional data processing and inference cost, compared to masked reconstruction pretraining. Although we minimize this cost by storing the teacher representations, it introduces additional hard disk and I/O burdens.

%%%%%%%%%%%%%%%%%%%%%%%%%%%%%%%%%%%%%%%%%%%%%%%%%%%%%%%%%%%%%%%%%%%%%%%%%%%%%%%
%%%%%%%%%%%%%%%%%%%%%%%%%%%%%%%%%%%%%%%%%%%%%%%%%%%%%%%%%%%%%%%%%%%%%%%%%%%%%%%

\end{document}